\documentclass[twoside,11pt]{article}

\usepackage{jmlr2e}
\usepackage[cmex10]{amsmath}
\usepackage{amssymb}
\usepackage{algorithm}
\usepackage{algorithmic}
\usepackage{graphicx}
\usepackage{epsfig}
\usepackage{subfigure}
\usepackage{booktabs}
\usepackage{multirow}

\jmlrheading{xx}{2012}{xx-xx}{xx/xx}{xx/xx}{N. Guan, D. Tao, Z. Luo, and J. Shawe-Taylor}

\ShortHeadings{MahNMF: Manhattan Non-negative Matrix Factorization}{N. Guan, D. Tao, Z. Luo, and J. Shawe-Taylor}
\firstpageno{1}

\begin{document}

\title{MahNMF: Manhattan Non-negative Matrix Factorization}

\author{\name Naiyang Guan \email Guan.Naiyang@uts.edu.au \\
       \addr Center for Quantum Computation and Intelligent Systems\\
       Faculty of Engineering and Information Technology\\
       University of Technology, Sydney\\
       Sydney, NSW 2007, Australia
       \AND
       \name Dacheng Tao \email dacheng.tao@uts.edu.au \\
       \addr Center for Quantum Computation and Intelligent Systems \\
       Faculty of Engineering and Information Technology \\
       University of Technology, Sydney\\
       Sydney, NSW 2007, Australia
       \AND
       \name Zhigang Luo \email zgluo@nudt.edu.cn \\
       \addr School of Computer Science \\
       National University of Defense Technology \\
       Changsha, Hunan 410073, China
       \AND
       \name John Shawe-Taylor \email J.Shawe-Taylor@cs.ucl.ac.uk \\
       \addr Centre for Computational Statistics and Machine Learning (CSML) \\
       Department of Computer Science \\
       University College London \\
       Gower Street, London WC1E 6BT, United Kingdom
       }

\editor{xx}

\maketitle

\begin{abstract}%   <- trailing '%' for backward compatibility of .sty file
Non-negative matrix factorization (NMF) approximates a non-negative matrix $X$ by a product of two non-negative low-rank factor matrices $W$ and $H$. NMF and its extensions minimize either the Kullback-Leibler divergence or the Euclidean distance between $X$ and $W^T H$ to model the Poisson noise or the Gaussian noise. In practice, when the noise distribution is heavy tailed, they cannot perform well. This paper presents Manhattan NMF (MahNMF) which minimizes the Manhattan distance between $X$ and $W^T H$ for modeling the heavy tailed Laplacian noise. Similar to sparse and low-rank matrix decompositions, e.g. robust principal component analysis (RPCA) and GoDec, MahNMF robustly estimates the low-rank part and the sparse part of a non-negative matrix and thus performs effectively when data are contaminated by outliers. We extend MahNMF for various practical applications by developing box-constrained MahNMF, manifold regularized MahNMF, group sparse MahNMF, elastic net inducing MahNMF, and symmetric MahNMF.

The major contribution of this paper lies in two fast optimization algorithms for MahNMF and its extensions: the rank-one residual iteration (RRI) method and Nesterov's smoothing method. In particular, by approximating the residual matrix by the outer product of one row of W and one row of $H$ in MahNMF, we develop an RRI method to iteratively update each variable of $W$ and $H$ in a closed form solution. Although RRI is efficient for small scale MahNMF and some of its extensions, it is neither scalable to large scale matrices nor flexible enough to optimize all MahNMF extensions. Since the objective functions of MahNMF and its extensions are neither convex nor smooth, we apply Nesterov's smoothing method to recursively optimize one factor matrix with another matrix fixed. By setting the smoothing parameter inversely proportional to the iteration number, we improve the approximation accuracy iteratively for both MahNMF and its extensions.

We conduct experiments on both synthetic and real-world datasets, such as face images, natural scene images, surveillance videos and multi-model datasets, to show the efficiency of the proposed Nesterov's smoothing method-based algorithm for solving MahNMF and its variants, and the effectiveness of MahNMF and its variants, by comparing them with traditional NMF, RPCA, and GoDec.
\end{abstract}

\begin{keywords}
  Non-negative Matrix Factorization (NMF), Nesterov's Smoothing Method, Sparse and Low-rank Matrix Decomposition
\end{keywords}

\section{Introduction}

Non-negative matrix factorization (NMF) is a popular matrix factorization approach that approximates a non-negative matrix $X$ by the product of two non-negative low-rank factor matrices $W$ and $H$. Different to other matrix factorization approaches, NMF takes into account the fact that most types of real-world data, particularly all images or videos, are non-negative and maintain such non-negativity constraints in factorization. This non-negativity constraint helps to learn parts-based representation supported by psychological and physical evidence \citep{Logothetis1996}\citep{Wachsmuth1994}. Therefore, NMF achieves great success in many fields such as image analysis \citep{Monga2007}, face recognition \citep{Zhang2008}, video processing \citep{Bucak2007}, and environmental science \citep{Paatero1994}.

NMF was first proposed by Paatero and Tapper \citep{Paatero1994} and was greatly popularized by Lee and Seung \citep{Lee1999}. Since then, many NMF variants have been proposed and have achieved great success in a variety of tasks. For example, Hoyer \citep{Hoyer2004} proposed sparseness constrained NMF (NMFsc) to enhance the sparseness of the learned factor matrices for computer vision tasks. Zafeiriou \textit{et al}. \citep{Zafeiriou2006} proposed discriminant NMF (DNMF) to incorporate Fisher's criteria for classification. Cai \textit{et al}. \citep{Cai2011} proposed graph regularized NMF (GNMF) to incorporate the geometric structure of a dataset for clustering. Recently, Sandler and Lindenbaum \citep{Sandler2011} proposed an earth mover's distance metric-based NMF (EMD-NMF) to model the distortion of images for several vision tasks. Liu \textit{et al}. \citep{Liu2012} proposed a constrained NMF (CNMF) to incorporate the label information as additional constraints for image representation.

From the mathematical viewpoint, traditional NMF \citep{Lee1999}\citep{Lee2001} and its variants minimize the Kullback-Leibler divergence and the Euclidean distance between $X$ and $W^T H$ to model the Poisson noise and Gaussian noise, respectively. Here, we call them KLNMF and EucNMF for short. Both KLNMF and EucNMF are popular because they can be efficiently optimized by using the multiplicative update rule \citep{Lee2001}. However, the noise in many practical applications is heavy tailed, so it cannot be well modeled by either Poisson distribution or Gaussian distribution. For example, the gradient-based image features such as SIFT \citep{Lowe2004} contain non-Gaussian heavy tailed noise \citep{Jia2011}. In these cases, traditional NMF does not perform well because it is not robust to outliers such as occlusions, Laplace noise, and salt \& pepper noise, whose distribution is heavy tailed.

On the other hand, real-world data often lies in a lower-dimensional subspace; for example, Basri and Jacobs \citep{Basri2003} showed that images taken from convex and Lambertian objects under distant illumination lie near an approximately nine-dimensional linear subspace. Recently, robust principal component analysis (RPCA, \citep{Candes2011}) and GoDec \citep{Zhou2011} have been proposed to robustly recover the lower-dimensional space in the presence of outliers. Both RPCA and GoDec consider the prior knowledge that noise, e.g., illumination/shadow in images and moving objects in videos, is sparse, and thus perform robustly in practice. Traditional NMF cannot robustly estimate the low-rank part of the data contaminated by outliers because it does not consider such prior knowledge of the sparse structure of noise.

In this paper, we present Manhattan NMF (MahNMF) to robustly estimate the low-rank part and the sparse part of a non-negative matrix. MahNMF models the heavy tailed Laplacian noise by minimizing the Manhattan distance between an $m\times n$-dimensional non-negative matrix $X$ and $W^T H$, i.e.,
\begin{equation}\label{eq1.1}
    \min_{W\ge 0,H\ge 0} f(W,H)=\|X-W^T H\|_M,
\end{equation}
where $\|\cdot\|_M$ is the Manhattan distance and the reduced dimensionality $r$ satisfies that $r\ll \min (m,n)$. Since both $W$ and $H$ are low-rank, MahNMF actually estimates the non-negative low-rank part, i.e., $W^T H$, and the sparse part, i.e., $X-W^T H$, of a non-negative matrix $X$. Benefiting from both the modeling capability of Laplace distribution to the heavy tailed behavior of noise and the robust recovery capability of the sparse and low-rank decomposition, such as RPCA and GoDec, MahNMF performs effectively and robustly when data are contaminated by outliers. We further extend MahNMF for various practical applications by developing box-constrained MahNMF, manifold regularized MahNMF, and group sparse MahNMF. These extensions follow the regularization theory by integrating MahNMF with various regularizations. By taking into account the grouping effect of the sparse part, we develop the elastic net inducing MahNMF to learn the low-rank and group sparse decomposition of a non-negative matrix. Inspired by spectral clustering, we develop a symmetric MahNMF for image segmentation. Although \citep{Lam2008} tried to model Laplacian noise in NMF, it cannot be used in practice because the semi-definite programming-based optimization method used suffers from both slow convergence and non-scalable problems.

The main contribution of this paper lies in two fast optimization methods for MahNMF and its extensions: the rank-one residual iteration (RRI) method and Nesterov's smoothing method. In particular, RRI approximates the residual matrix with the outer product of one row of $W$ and one row of $H$ in \eqref{eq1.1} and iteratively updates each variable of $W$ and $H$ in a closed form solution. RRI is efficient for optimizing small-scale MahNMF and some of its extensions, but it is neither scalable to large scale matrices nor flexible enough to optimize all MahNMF extensions. Since the objective functions of MahNMF and its extensions are neither convex nor smooth, we apply Nesterov's smoothing method to recursively optimize one factor matrix with another matrix fixed. By setting the smoothing parameter inversely proportional to the iteration number, we improve the approximation accuracy iteratively for both MahNMF and its extensions.

We conduct experiments on both synthetic and real-world datasets, such as face images, natural scene images, surveillance videos and multi-model datasets, to show the efficiency of the proposed Nesterov's smoothing method-based algorithms for optimizing MahNMF and its variants and their effectiveness in face recognition, image clustering, background/illumination modeling, and multi-view learning by comparing them with traditional NMF, RPCA, and GoDec.

The remainder of this paper is organized as follows: Section II presents the rank-one residual iteration (RRI) method for optimizing MahNMF, and Section III presents Nesterov's smoothing method. Section IV presents several MahNMF extensions which can be solved by using the proposed Nesterov smoothing method-based algorithm. In Section V, we conduct experiments to show both the efficiency of Nesterov smoothing method-based algorithm for MahNMF and the effectiveness of MahNMF and its variants. Section VI concludes this paper.

\textit{Notations}: We denote by a lower-case $x$, a headed $\vec{x}$ and a capital $X$ a scalar, vector and matrix, respectively. In particular, $\vec{1}$ signifies a vector full of one, $I$ signifies an identity matrix, and $0$ signifies zero, zero vector or null matrix. We denote by bracketed subscript and superscript the elements of a vector or a matrix, e.g., $\vec{x}_{(k)}$ signifies the $k$-th element of $\vec{x}$, and $X_{(k)}$, $X^{(k)}$, $X_{(i,j)}$ signify the $k$-th row, the $k$-th column, and the $(i,j)$-th element of $X$, respectively. We denote by a subscript, e.g., $\vec{x}_k$ and $X_k$ the points in a sequence. We denote by $\mathrm{R}$ and $\mathrm{R}_+$ the set of real numbers and the set of non-negative real numbers, respectively. Consequently, $\mathrm{R}_+^m$ and $\mathrm{R}_+^{m\times n}$ signify the set of $m$-dimensional non-negative vectors and the set of $m\times n$-dimensional non-negative matrices, respectively. We denote by $\|\vec{x}\|_{l_1}$ and $\|\vec{x}\|_{l_2}$ the $l_1$ and $l_2$ norm of a vector $\vec{x}$, respectively. For any matrices $X\in \mathrm{R}^{m\times r}$ and $Y\in \mathrm{R}^{m\times r}$, we denote their Euclidean distance (Frobenius norm) and Manhattan distance as $\|X-Y\|_F$ and $\|X-Y\|_M$, respectively. In addition, we denote by $X\circ Y$  and $\frac{X}{Y}$ their element-wise product and division, respectively.

\section{Rank-one Residual Iteration Method for MahNMF}
Since the objective function \eqref{eq1.1} is non-convex, we recursively optimize one factor matrix $W$ or $H$ with another fixed, i.e., at iteration $t\ge 0$, we update
\begin{equation}
    H_{t+1} = {\arg\min}_{H\ge 0} \|X-W_t^T H\|_M, \label{eq2.1}
\end{equation}
and
\begin{equation}
    W_{t+1} = {\arg\min}_{W\ge 0} \|X^T-H_{t+1}^T W\|_M, \label{eq2.2}
\end{equation}
until convergence. The convergence is usually checked by the following objective-based stopping condition:
\begin{equation}
    |f(W_t,H_t)-f(W_{t+1},H_{t+1})|\le \xi, \label{eq2.3}
\end{equation}
where $\xi$ is the precision, e.g., $\xi=.1$. Because problems \eqref{eq2.1} and \eqref{eq2.2} are symmetric, we focus on optimizing \eqref{eq2.1} in the following section, and \eqref{eq2.2} can be solved in a similar way.

Although \eqref{eq2.1} is convex, the Manhattan distance-based objective function, i.e., $f(W_t,H)$, is non-differentiable when $X-W_t^T H$ contains zero elements. This means that the gradient-based method cannot be directly applied to optimizing \eqref{eq2.1}. Fortunately, we will show that each variable in $H$ can be updated in a closed form solution and thus \eqref{eq2.1} can be optimized by using alternating optimization over each variable of $H$. Given $W^T$ and rows of $H$ except $H_{(l)}$, eq. \eqref{eq2.1} can be written as
\begin{equation}
    \min_{H_{(l)}\ge 0} \|Z-W_{(l)}^T H_{(l)}\|_M, \label{eq2.4}
\end{equation}
where $Z=X-\sum_{i=1,i\ne l}^r W_{(i)}^T H_{(i)}$ is the residual matrix. Actually, Eq. \eqref{eq2.4} is a rank one approximation of the residual matrix. Therefore, following \citep{Ho2011}, we term this method the rank-one residual iteration (RRI) method.

\begin{figure*}[ht]
\centering
\includegraphics[width=0.5\linewidth]{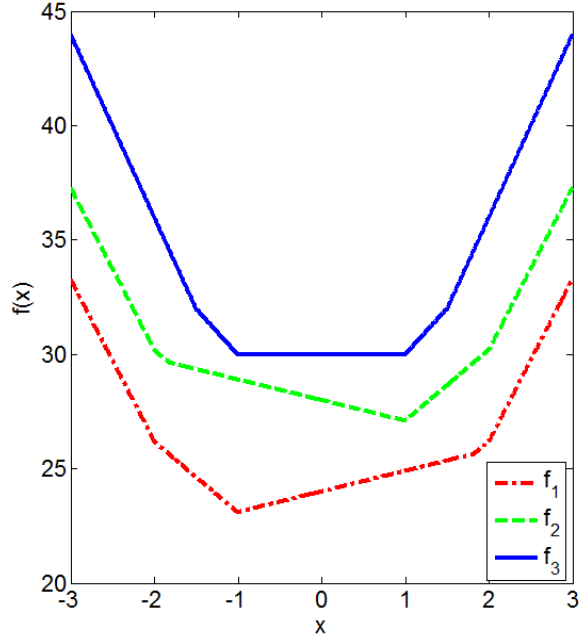}
\caption{Piecewise function examples. The minimums of $f_1$ and $f_2$ appear at $-1$ and $1$, respectively, while the minimum of $f_3$ appears at the range of $[-1,1]$.}
\label{fig.piecewise-func}
\end{figure*}

Since \eqref{eq2.4} is convex and separable with respect to each variable $H_{(l,j)}$, wherein $j\in \{1,...,n\}$, there exists the optimal solution and $H_{(l,j)}$ is updated as follows
\begin{align}
    \min_{H_{(l,j)}\ge 0} \|Z^{(j)}-W_{(l)}^T H_{(l,j)}\|_1 &= |W_{(l,1)} H_{(l,j)}-Z_{(1,j)}|+...+|W_{(l,m)} H_{(l,j)}-Z_{(m,j)}| \nonumber \\
        &\triangleq \zeta_{(l,j)} (H_{(l,j)}). \label{eq2.5}
\end{align}
Looking carefully at $\zeta_{(l,j)} (H_{(l,j)})$, it is a continuous piecewise linear function whose piecewise points are $ \mathbf{P}=\{p_s=\frac{Z_{(i_s,j)}}{W_{(l,i_s)}}|W_{(l,i_s)}\ne 0,i_s\in \{1,...,m\},s=1,...q,q\le m\}$. It is obvious that the minimum of $\zeta_{(l,j)} (H_{(l,j)})$ appears at one point of $\mathbf{P}$. Regardless of the constraint $H_{(l,j)}\ge 0$, the point that first changes the sign of the slope of $\zeta_{(l,j)} (H_{(l,j)})$ is its minimum. By sorting $\mathbf{P}$ in an ascending order, we have $p_{s^1}\le\cdots\le p_{s^c}\le\cdots\le p_{s^q}$, wherein $s^c\in \{1,...,q\}$. Furthermore, by sorting $\{W_{(l,i_{s^c})},c=1,...,q\}$ accordingly, we can remove the absolute operator in \eqref{eq2.5} and rewrite it into $q+1$ pieces as follows
\begin{equation}
    \zeta_{(l,j)}(x)=\left \{\begin{array}{c@{\;}l}
    (-W_{(l,i_{s^1})}-...-W_{(l,i_{s^q})})x+Z_{(i_{s^1},j)}+...+Z_{(i_{s^q},j)},& x\le p_{s^1} \\
    (W_{(l,i_{s^1})}-...-W_{(l,i_{s^q})})x-Z_{(i_{s^1},j)}+...+Z_{(i_{s^q},j)},& p_{s^1}\le x\le p_{s^2} \\
    (W_{(l,i_{s^1})}+...-W_{(l,i_{s^q})})x-Z_{(i_{s^1},j)}-...+Z_{(i_{s^q},j)},& p_{s^{q-1}}\le x\le p_{s^q} \\
    (W_{(l,i_{s^1})}+...+W_{(l,i_{s^q})})x-Z_{(i_{s^1},j)}-...-Z_{(i_{s^q},j)},& p_{s^q}\le x
    \end{array} \right. \label{eq2.6}
\end{equation}
Since $W_{(l,i_{s^c})}>0$, the slope of the piecewise function in \eqref{eq2.6} is increasing. It is easy to find the point which first changes the sign of slope. Suppose $p_{s^c}$ first changes the signs of slope of $\zeta_{(l,j)}(x)$, i.e., $W_{(l,i_{s^1})}+...-W_{(l,i_{s^c})}-...-W_{(l,i_{s^q})}<0$ and $W_{(l,i_{s^1})}+...+W_{(l,i_{s^c})}-...-W_{(l,i_{s^q})}\ge 0$. It is clear that $p_{s^c}$ minimizes $\zeta_{(l,j)}(H_{(l,j)})$. Note that the minimum is not unique because the slope at $p_{s^c}$ may be zero. See Figure \ref{fig.piecewise-func} for three examples of piecewise functions; it is clear that $-1$ and $1$ minimizes $f_1$ and $f_2$ and any point in the range $[-1,1]$ minimizes $f_3$. Taking into account the non-negativity constraint, we obtain the solution of \eqref{eq2.5} as $\max\{0,p_{s^c}\}$. In the case of $f_3$ in Figure \ref{fig.piecewise-func}, we simply take the leftmost point $p_{s^c}$ as its optimal solution.

We summarize the RRI method in \textit{Algorithm} \ref{alg.rri}. It successively updates each row of $H$ and stops when the following stopping condition is satisfied
\begin{equation}
    |f(W,H_{k+1})-f(W,H_k)|\le \epsilon, \label{eq2.7}
\end{equation}
where $\epsilon$ is the precision, e.g., $\epsilon=.1$. By recursively solving \eqref{eq2.1} and \eqref{eq2.2} with \textit{Algorithm} \ref{alg.rri}, the MahNMF problem \eqref{eq1.1} can be successfully solved. The previous variable is used as a warm start, i.e., $H_0=H_t^*$, to accelerate convergence of the RRI method.

\begin{algorithm}
\caption{RRI Method for MahNMF}
\label{alg.rri}
\begin{algorithmic}
\STATE \textbf{Input}: $X\in \mathrm{R}_+^{m\times n}, W\in \mathrm{R}_+^{r\times m}, H_0\in \mathrm{R}_+^{r\times n}$.
\STATE \textbf{Output}: $H_{t+1}^*$.
\STATE 1: Initialize $W_l^{'}=[W_{(l,i_1)},...,W_{(l,i_{q_l})}]$, $l\in {1,...,r}$, $q_c\in {1,...,m}$.
\STATE 2: For $k=0,1,2,...$
\STATE 3: \quad For $l=1,...,r$
\STATE 4: \quad\quad Compute $Z=X-\sum_{i=1,i\ne l}^r W_{(i)}^T {H_k}_{(i)}$.
\STATE 5: \quad\quad Compute $\mathbf{P}_j=\{p_s=\frac{Z_{(i_s,j)}}{W_{(l,i_s)}} |W_{(l,i_s)}\ne 0,i_s\in \{1,...,m\},s=1,...q\}$, for $j=1,...,n$.
\STATE 6: \quad\quad Sort $\mathbf{P}_j$ simultaneously and sort $W_l^{(j)}$ according to $\mathbf{P}_j$'s order, for $j=1,...,n$.
\STATE 7: \quad\quad Find the piecewise point that first changes the sign of $\zeta_l (x)$ in \eqref{eq2.6}.
\STATE 8: \quad\quad Update ${H_{k+1}}_{(l,j)}=\max \{0,p_{s^{c_j}}\}$ simultaneously for $j=1,...,n$.
\STATE 9: \quad End For
\STATE 10: \; Check the stopping condition \eqref{eq2.7}.
\STATE 11: End For
\STATE 12: $H_{t+1}^*=H_{k+1}$.
\end{algorithmic}
\end{algorithm}

The main time cost of \textit{Algorithm} \ref{alg.rri} is spent on sentence 6 that sorts the piecewise points. Its time complexity is $O(nmlogm)$ because the sorting operator for each piecewise point set costs $O(mlogm)$ time in the worst case. Sentence 7 finds the piecewise point that first changes the sign of $\zeta_{(l,j)}(x)$ from the sorted set. This can be done by initializing the slope as $-\sum_{c=1}^q W_{(l,i_{s^c})}$  and increasing it by $2W_{(l,i_{s^c})}$ at the $c$-th step. Once the sign of slope changes, the procedure stops and outputs $p_{s^c}$. The worst case time complexity of this procedure is $O(mn)$. Therefore, the total complexity of \textit{Algorithm} \ref{alg.rri} is $O(mnr(\log m+1))\times K$, where $K$ is the iteration number. Since \textit{Algorithm} \ref{alg.rri} finds the closed form solution for each variable of $H$, it converges fast. However, the time complexity is high especially when $m$ is large. Therefore, RRI is not scalable for large scale problems. In the following section, we propose an efficient and scalable algorithm for optimizing MahNMF.

\section{Nesterov's Smoothing Method for MahNMF}
Since the Manhattan distance equals the summation of the $l_1$ norm, i.e., $\|X-W_t^T H\|_M=\sum_{j=1}^n \|X^{(j)}-W_t^T H^{(j)}\|_{l_1}$, the minimization problem \eqref{eq2.1} can be solved by optimizing each column of H separately. For the $j$-th column, the sub-problem is
\begin{equation}
    \min_{H^{(j)}\ge 0} \|X^{(j)}-W_t^T H^{(j)}\|_{l_1}. \label{eq3.1}
\end{equation}
Without the non-negativity constraint, eq. \eqref{eq3.1} shrinks to the well-known least absolute deviations (LAD, \citep{Karst1958}) regression problem. Here, we term \eqref{eq3.1} as a non-negative LAD (NLAD) problem for the convenience of presentation. According to \citep{Harter1974}, LAD is much more robust than the least squares (LS) method especially on the datasets contaminated by outliers. NLAD inherits the robustness from LAD and keeps the non-negativity capability of datasets.

For any given observation $\vec{x}=X^{(j)}\in \mathrm{R}_+^m, 1\le j\le n$ and a matrix $W=W_t\in \mathrm{R}_+^{r\times m}$, the NLAD problem \eqref{eq3.1} can be written as
\begin{equation}
    \min_{\vec{h}} \{ f(W,\vec{h})=\|W^T \vec{h}-\vec{x}\|_{l_1}=\sum_{i=1}^m |<W^{(i)},\vec{h}>_1-\vec{x}_{(i)}|: \vec{h}\in Q_1=\mathrm{R}_+^r\}, \label{eq3.2}
\end{equation}
where $<\cdot,\cdot>_1$ signifies the inner product in $\mathrm{R}^r$. Define the norm that endows the domain $E_1=\mathrm{R}^r$ as $\|\vec{h}\|_1=\|\vec{h}\|_{l_2}=(\sum_{j=1}^r \vec{h}_{(j)}^2)^{\frac{1}{2}}$, and construct the prox-function for the feasible set $Q_1$ as $d_1(\vec{h})=\frac{1}{2} \|\vec{h}\|_1^2$. It is obvious that $d_1(\cdot)$ is strongly convex and the convexity parameter is $\delta_1=1$. Since $E_1$ is a self-dual space, we know that the dual norm $\|\vec{w}\|_1^*=\max_{\vec{y}} \{<\vec{w},\vec{y}>_1: \vec{y}\in E_1,\|\vec{y}_1\|=1\}=\|\vec{w}\|_1$ for any $\vec{w}\in E_1$.

Since $f(W,\vec{h})$ is convex and continuous, there must be an optimal solution. However, it cannot be solved directly by using the gradient-based method because $f(W,\vec{h})$ is non-smooth. Fortunately, Nesterov \citep{Nesterov2004} shows that $f(W,\vec{h})$ can be approximated by a smooth function. In particular, we first construct a dual function for the primal non-smooth function and smooth the dual function by adding a smooth and strongly convex prox-function for the feasible set of the dual variable. Then we solve the smoothed dual function in the dual space and project the solution back to primal space. The obtained solution can be considered as an approximate minimum of the primal function. By choosing the dual domain $E_2=\mathrm{R}^m$ and the feasible set $Q_2=\{\vec{\mu}\in E_2: |\vec{\mu}_{(i)}|\le 1,i=1,...,m\}$, wherein $\vec{\mu}$ is the dual variable, the primal problem \eqref{eq3.2} is equivalently rewritten as
\begin{equation*}
    \min_{\vec{h}} \{f(W,\vec{h})=\max_{\vec{\mu}} \{<W^T\vec{h}-\vec{x},\vec{\mu}>_2: \vec{\mu}\in Q_2\}: \vec{h}\in Q_1\},
\end{equation*}
where $<\cdot,\cdot>_2$ is the inner product in $\mathrm{R}^m$. The corresponding dual problem is
\begin{equation*}
    \max_{\vec{\mu}} \{\phi(\vec{\mu})=\min_{\vec{h}}\{<W^T\vec{h}-\vec{x},\vec{\mu}>_2: \vec{h}\in Q_1\}: \vec{\mu}\in Q_2\}.
\end{equation*}

Since $\vec{v}\approx W\vec{h}$, $Q_1$ is bounded, i.e., there exists a positive number $M_1$ such that $\vec{h}_{(j)}\le M_1$ for any $\vec{h}\in Q_1$. Then the dual function $\phi(\vec{\mu})$ can be calculated explicitly, i.e., $\phi(\vec{\mu })=<W^T \varphi_{M_1} (W\vec{\mu})-\vec{x},\vec{\mu}>_2$, wherein $\varphi_{M_1} (\cdot)$ is an element-wise operator defined as $\varphi_{M_1}(a)=\left \{\begin{array}{c@{\;}l} 0,& a\ge 0\\ M_1,& a<0 \end{array} \right.$. Since it is difficult to estimate $M_1$, the dual problem is still difficult to solve. However, it can be easily solved by adding a simple prox-function. According to \citep{Nesterov2004}, we define the prox-function for $Q_2$ as $d_2 (\vec{\mu}=\frac{1}{2} \|\mu\|_2^2=\frac{1}{2}(\sum_{i=1}^m \|W^{(i)}\|_1^* \vec{\mu}_{(i)}^2)^{\frac{1}{2}}$. By adding the prox-function, we obtain a smoothed approximate function for $f(W,\vec{h})$ as follows
\begin{align}
    f_{\lambda} (W,\vec{h}) &= \max_{\vec{\mu}} \{<W^T\vec{h}-\vec{x},\vec{\mu}>_2-\lambda d_2(\vec{\mu}):\vec{\mu}\in Q_2\} \nonumber \\
        &= \max_{\vec{\mu}} \{\sum_{i=1}^m ({W^{(i)}}^T \vec{h}-\vec{x}_{(i)})\vec{\mu}_{(i)}-\frac{1}{2} \lambda \sum_{i=1}^m \|W^{(i)}\|_1^*\vec{\mu}_{(i)}^2:\vec{\mu}\in Q_2\}, \label{eq3.3}
\end{align}
where $\lambda>0$ is a parameter that controls the smoothness. The larger the parameter $\lambda$, the smoother the approximate function $f_\lambda (W,\vec{h})$ and the worse its approximate accuracy. Using algebra, eq. \eqref{eq3.3} can be written as
\begin{equation}
    f_\lambda (W,\vec{h}) = \max_{\vec{\mu}} \{(W^T \vec{h}-\vec{x})^T \vec{\mu}-\frac{1}{2} \vec{\mu}^T A\vec{\mu}:|\vec{\mu}_{(i)}|\le 1\}, \label{eq3.4}
\end{equation}
where
\begin{equation*}
    A = \left[ \begin{array}{ccc} \lambda \|W^{(1)}\|_1^* & \cdots & 0 \\
    \vdots & \ddots & \vdots \\
    0 & \cdots & \lambda \|W^{(m)}\|_1^* \end{array} \right].
\end{equation*}
Let $\rho\in \mathrm{R}^m$ and $\varphi\in \mathrm{R}^m$ to be the Lagrange multiplier vectors corresponding to the constraints, i.e., $-\vec{\mu}_{(i)}-1\le 0$ and $\vec{\mu}_{(i)}-1\le 0$, respectively, the K.K.T. conditions of \eqref{eq3.4} are as follows
\begin{equation}
    \left \{ \begin{array}{c}
    W\vec{h}-\vec{x}-A\vec{\mu}-\vec{\rho}+\vec{\varphi}=0 \\
    \rho_{(i)}\ge 0,\quad \varphi_{(i)}\ge 0 \\
    -\vec{\mu}_{(i)}-1\le 0,\quad \vec{\mu}_{(i)}-1\le 0 \\
    (-\vec{\mu}_{(i)}-1)\vec{\rho}_{(i)}=0,\quad (\vec{\mu}_{(i)}-1)\vec{\varphi}_{(i)}=0
    \end{array} \right.. \label{eq3.5}
\end{equation}
From \eqref{eq3.5}, we can easily obtain the closed-form solution of \eqref{eq3.4} as
\begin{equation}
    \vec{\mu}_{(i)}^*=med \{1,-1,\frac{{W^{(i)}}^T \vec{h}-\vec{x}_{(i)}}{\lambda \|W^{(i)}\|_1^*}\},\; i=1,...,m,
\end{equation}
where $med(\cdot)$ is the median operator. By substituting $\vec{\mu}^*$ back into \eqref{eq3.4}, we obtain the closed-form smoothed function $f_\lambda (W,\vec{h})$ as
\begin{equation}
    f_\lambda (W,\vec{h}) = \sum_{i=1}^m \|W^{(i)}\|_1^* \psi_\lambda (\frac{|{W^{(i)}}^T \vec{h}-\vec{x}_{(i)}|}{\|W^{(i)}\|_1^*}), \label{eq3.6}
\end{equation}
where $\psi_\lambda (\tau)=\left \{ \begin{array}{c@{\;}l} \frac{\tau^2}{2\lambda},&0\le\tau\le\lambda\\ \tau-\frac{\lambda}{2}, &\tau\ge\lambda \end{array} \right.$. According to Theorem 1 in \citep{Nesterov2004}, $f_\lambda (W,\vec{h})$ is well defined and continuously differentiable at any $\vec{h}\in E_1$. Moreover, $f_\lambda (W,\vec{h})$ is convex and its gradient $\nabla f_\lambda (W,\vec{h})=W\vec{\mu}^*$ is Lipschitz continuous with constant $L_\lambda=\frac{1}{\lambda\delta_2} \|W^T\|_{1,2}^2$, wherein $\delta_2=1$ is the convexity parameter of $d_2(\cdot)$ and $\|W^T\|_{1,2}$ is the norm of projection matrix $W$ which is defined as follows
\begin{align*}
    \|W^T\|_{1,2} &= \max_{\vec{h},\vec{\mu}} \{\sum_{i=1}^m \vec{\mu}_{(i)}<W^{(i)},\vec{h}>_1: \|\vec{h}\|_1\le 1,\|\vec{\mu}\|_2\le 1\} \nonumber \\
    &\le \max_{\vec{\mu}} \{\sum_{i=1}^m \|W^{(i)}\|_1^* \vec{\mu}_{(i)}: \sum_{i=1}^m \|W^{(i)}\|_1^* \vec{\mu}_{(i)}^2\le 1\}=[\sum_{i=1}^m \|W^{(i)}\|_1^*]^{\frac{1}{2}} \triangleq D^{\frac{1}{2}}.
\end{align*}
By using the obtained smoothed function, \eqref{eq3.1} can be approximately solved by
\begin{equation}
    H^{(j)} = {\arg\min}_{\vec{h}\ge 0} f_\lambda (W_t,\vec{h}). \label{eq3.7}
\end{equation}

Since $f_\lambda(W,\vec{h})$ is smooth, convex and its gradient is Lipschitz continuous, it naturally motivates us to optimize \eqref{eq3.7} by using Nesterov's optimal gradient method (OGM, \citep{Nesterov2004}). In particular, OGM constructs two auxiliary sequences in optimization: one sequence stores the historical gradients and another sequence stores the search point that minimizes the quadratic approximation of $f_\lambda (W,\vec{h})$ at the current solution. The step size is determined by the Lipchitz constant. In each iteration round, the solution is updated by combining the historical gradients and search point. This combination accelerates the gradient method and makes OGM achieve an optimal convergence rate of $O(\frac{1}{k^2})$ for optimizing \eqref{eq3.7}. In this paper, the search points $\{\vec{y}_k\}$ and the historical gradients $\{\vec{z}_k\}$ are defined as follows:
\begin{equation}
    \vec{y}_k = {\arg\min}_{\vec{y}\in Q_1} \{<\nabla f_\lambda (W,\vec{h}_k),\vec{y}-\vec{h}_k>_1+\frac{L_\lambda}{2}\|\vec{y}-\vec{h}_k\|_1^2\}, \label{eq3.8.1}
\end{equation}
and
\begin{equation}
    \vec{z}_k = {\arg\min}_{\vec{z}\in Q_1} \{\frac{L_\lambda}{\delta_1} d_1(\vec{z})+\sum_{i=0}^k \frac{i+1}{2} [f_\lambda (W,\vec{h}_i)+<\nabla f_\lambda (W,\vec{h}_i),\vec{z}-\vec{h}_i>_1]\}, \label{eq3.8.2}
\end{equation}
where $k\ge 0$ is the iteration counter. By solving \eqref{eq3.8.1} and \eqref{eq3.8.2}, respectively, we have
\begin{equation}
    \vec{y}_k = \max (0,\vec{h}_k-\frac{1}{L_\lambda}\nabla f_\lambda (W,\vec{h}_k)), \label{eq3.9.1}
\end{equation}
and
\begin{equation}
    \vec{z}_k = \max (0,-\frac{1}{L_\lambda} \sum_{i=0}^k \frac{i+1}{2} \nabla f_\lambda (W,\vec{h}_i)). \label{eq3.9.2}
\end{equation}
According to \citep{Nesterov2004}, we combine $\vec{y}_k$ and $\vec{z}_k$ as follows:
\begin{equation}
    \vec{h}_{k+1} = \frac{2}{k+3}\vec{z}_k+\frac{k+1}{k+3}\vec{y}_k. \label{eq3.10}
\end{equation}
By alternating between \eqref{eq3.9.1}, \eqref{eq3.9.2} and \eqref{eq3.10} until convergence, we obtain the final solution $\vec{h}_\lambda^*$ of \eqref{eq3.7}. The convergence is checked by using the following objective-based stopping condition:
\begin{equation}
    |f_\lambda (W,\vec{h}_k)-f_\lambda (W,\vec{h}_\lambda^*)|\le \epsilon, \label{eq3.11}
\end{equation}
where $\epsilon$ is the precision, e.g., $\epsilon=.1$. Since $\vec{h}_\lambda^*$ is unknown in ahead, we usually use $\vec{h}_{k+1}$ instead. According to Theorem 3 of \citep{Nesterov2004}, the complexity of finding an $\epsilon$-solution does not exceed $N=4\|\frac{\|W^T\|_{1,2}}{\epsilon}\sqrt{\frac{D_1D_2}{\delta_1\delta_2}}+2\sqrt{\frac{MD_1}{\delta_1}}$. By substituting $\delta_1=\delta_2=1$, $M=0$, $D_2=\frac{D}{2}$, we have $N=\frac{2D\sqrt{2D_1}}{\epsilon}$, namely OGM finds an $\epsilon$-solution for \eqref{eq3.1} in $O(\frac{1}{\epsilon})$ iterations.

\begin{figure*}[ht]
\centering
\includegraphics[width=1.0\linewidth]{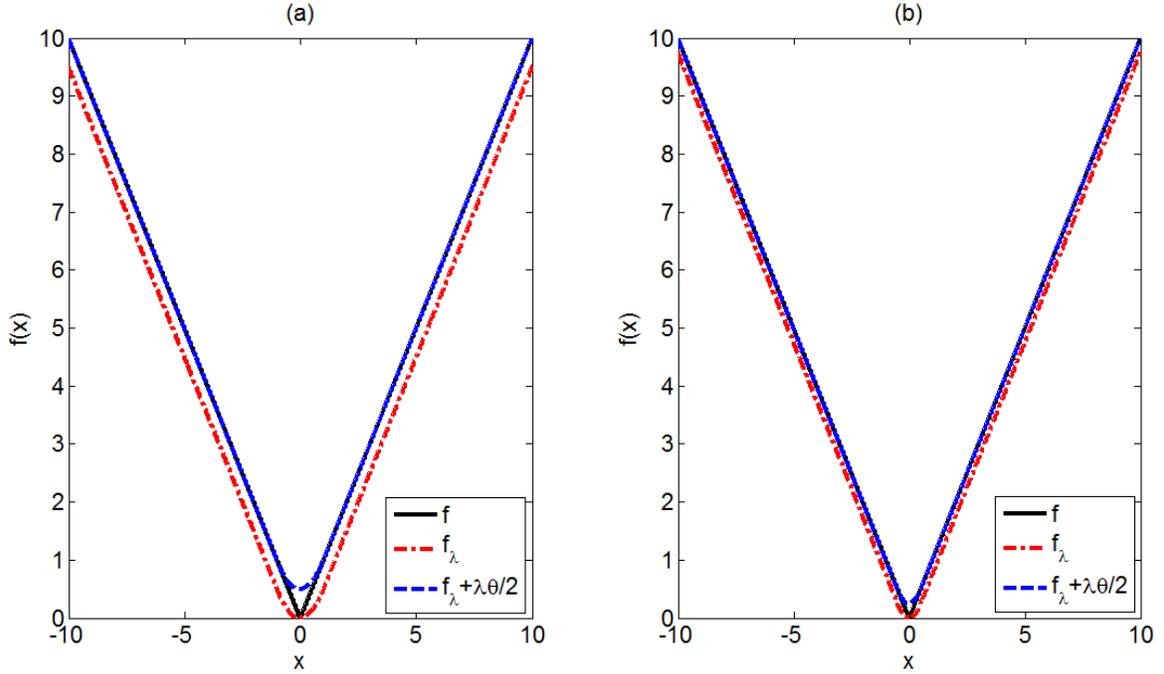}
\caption{Two examples of smoothing function of the absolute function f when (a) $\lambda=.1$ and (b) $\lambda=.05$ with $\theta=10$.}
\label{fig.smooth-func}
\end{figure*}

As mentioned above, the smooth parameter $\lambda$ controls the approximation of $f_\lambda (W,\vec{h})$ for $f(W,\vec{h})$, smaller $\lambda$ implies better approximation. A natural question is whether $\vec{h}_\lambda^*$ minimizes \eqref{eq3.1} as $\lambda$ goes to zero. To answer this question, we first show that $f(W,\vec{h})$ is bounded and gets infinitely close to $f_\lambda (W,\vec{h})$ as $\lambda$ goes to zero in the following \textit{Theorem 1} and we prove that the smoothing method finds an approximate solution of MahNMF in \textit{Theorem 2}. Figure \ref{fig.smooth-func} gives two examples of the smoothing functions with different smooth parameters. It shows that the original non-smooth function is bounded.

\noindent
{\bf Theorem 1} {\it Given any positive number $\lambda>0$, we have the following inequality:
\[
	f_\lambda (W,\vec{h})\le f(W,\vec{h})\le f_\lambda (W,\vec{h})+\frac{D}{2}\lambda.
\]}

\noindent
{\bf Proof}. Defining the residual error $\vec{e}=W^T \vec{h}-\vec{x}$, then its $i$-th entry is $\vec{e}_{(i)}={W^{(i)}}^T \vec{h}-\vec{x}_{(i)}$. The approximation function $f_\lambda (W,\vec{h})$ can be written as the following function with respect to $\vec{e}$:
\begin{equation}
    f_\lambda (W,\vec{h}) = \sum_{i=1}^m \|W^{(i)}\|_1^* \psi_\lambda (\frac{|\vec{e}_{(i)}|}{\|W^{(i)}\|_1^*}). \label{eq3.12}
\end{equation}
Below we will prove that $\|{W^{(i)}}^T\|_1^* \psi_\lambda (\frac{|\vec{e}_{(i)}|}{\|W^{(i)}\|_1^*}) \le |\vec{e}_{(i)}|\le \|W^{(i)}\|_1^* \psi_\lambda (\frac{|\vec{e}_{(i)}|}{\|W^{(i)}\|_1^*})+\frac{\lambda \|W^{(i)}\|_1^*}{2}$. For the convenience of derivation, we focus on the following function $g_{\lambda,\theta}(x)=\theta \psi_\lambda (\frac{x}{\theta})$, wherein $x\ge 0$. According to the definition of $\psi_\lambda(\cdot)$ in \eqref{eq3.6}, we have
\begin{equation}
    g_{\lambda,\theta}(x)-x=\left \{ \begin{array}{c@{\;}l}
    \frac{x^2}{2\lambda\theta}-x=x(\frac{x}{2\lambda\theta}-1)=\frac{1}{2\lambda\theta} (x-\lambda\theta)^2-\frac{\lambda\theta}{2},& 0\le x\le \lambda\theta \\
    -\frac{\lambda\theta}{2},& x\ge \lambda\theta \end{array} \right. \label{eq3.13}
\end{equation}
It is obvious that $g_{\lambda,\theta}(x)\le x \le g_{\lambda,\theta}(x)+\frac{\lambda\theta}{2}$. By substituting these inequalities into \eqref{eq3.13} and considering $D=\sum_{i=1}^m \|W^{(i)}\|_1^*$, we have $f_\lambda (W,\vec{h})\le f(W,\vec{h})\le f_\lambda (W,\vec{h})+\frac{D}{2}\lambda$. This completes the proof. \hfill\BlackBox

Since the columns of $H$ are separable, OGM can be written in a matrix form and summarized in \textit{Algorithm} \ref{alg.ogm}, wherein $Q=\left[\begin{array}{c} \|W^{(1)}\|_1^*\\ \vdots \\ \|W^{(m)}\|_1^* \end{array}\right]\times \vec{1}_n^T$ and $L_\lambda=\frac{1}{\lambda\delta_2}\|W^T\|_{1,2}^2$. \textit{Algorithm} \ref{alg.ogm} accepts the smooth parameter $\lambda$ as an input and outputs an approximate solution of the sub-problem \eqref{eq2.1}.

\begin{algorithm}
\caption{OGM for Smoothed NLAD}
\label{alg.ogm}
\begin{algorithmic}
\STATE \textbf{Input}: $X\in \mathrm{R}_+^{m\times n}$, $W\in \mathrm{R}_+^{r\times m}$, $H_0\in \mathrm{R}_+^{r\times n}$, $\lambda$, $\epsilon$.
\STATE \textbf{Output}: $H_{t+1}^*$.
\STATE 1: Initialize $Q$, $L_\lambda$.
\STATE 2: For $k=0,1,2,...$
\STATE 3: \quad Compute $U_k=med \{1,-1,\frac{W^T H_k-X}{\lambda Q}\}$.
\STATE 4: \quad Compute $\nabla f_¦Ë(W,H_k)=WU_k$.
\STATE 5: \quad Compute $Y_k=\max (0,H_k-\frac{1}{L_\lambda} \nabla f_\lambda (W,H_k))$.
\STATE 6: \quad Compute $Z_k=\max (0,-\frac{1}{L_\lambda} \sum_{i=0}^k \frac{i+1}{2} \nabla f_\lambda(W,H_k))$.
\STATE 7: \quad Update $H_{k+1}=\frac{2}{k+3} Z_k+\frac{k+1}{k+3} Y_k$.
\STATE 8: \quad Check the stopping condition \eqref{eq2.7}.
\STATE 9: End For
\STATE 10: $H_{t+1}^*=H_{k+1}$.
\end{algorithmic}
\end{algorithm}

According to \citep{Nesterov2004}, \textit{Algorithm} \ref{alg.ogm} converges at the rate of $O(\frac{1}{k^2})$ for optimizing \eqref{eq3.7} and needs $O(\frac{1}{\epsilon})$ iterations to yield an $\epsilon$-solution of the original problem \eqref{eq3.1}. Since the distance between the primal and dual functions is
\begin{equation}
    0\le f(W,\vec{y}_N)-\phi(\hat{\mu})\le \lambda D_2+\frac{4\|W^T\|_{1,2}^2 D_1}{\lambda\delta_1\delta_2(N+1)^2}\le \epsilon, \label{eq3.14}
\end{equation}
where $D_1=\max_{\vec{h}} \{d_1(\vec{h}): \vec{h}\in Q_1\}$, and $\hat{\mu}=\sum_{i=0}^N \frac{2(i+1)}{(N+1)(N+2)}\vec{\mu}_\lambda(\vec{h}_i)$, and $\vec{\mu}_\lambda(\vec{h}_i)$ is the solution of \eqref{eq3.10} at the $i$-th iteration rounds. By minimizing the right-hand side of the above inequality, we have $\lambda=\frac{\epsilon}{D_2}$ and $N+1\le 4\|W^T\|_{1,2}\sqrt{\frac{D_1D_2}{\delta_1\delta_2}}$. Since $\vec{x}\approx W^T \vec{h}$, $D_1$ is bounded. However, this bound is difficult to calculate exactly. In the following section, we will show that this deficiency can be overcome by slightly modifying the feasible set $Q_1$.

According to \citep{Nesterov2004}, sentence 5 of \textit{Algorithm} \ref{alg.ogm} can be slightly changed to guarantee decreasing the objective function. In particular, we find $Y_k^{'}=\max(0,H_k-\frac{1}{L_\lambda}\nabla f_\lambda(W,H_k))$ and set $Y_k={\arg\min}_Y \{f_\lambda(W,Y),Y\in \{Y_{k-1},H_k,Y_k^{'}\}\}$. This strategy requires additional computation of the objective function and thus increases the time cost of each iteration by $O(mn)$. The main time cost of \textit{Algorithm} \ref{alg.ogm} is spent on sentences 3 and 4 to calculate the gradient, whose complexity is $O(mnr)$. Therefore, the total time complexity of \textit{Algorithm} \ref{alg.ogm} is $O(2mn(r+1))\times K$, wherein $K$ is the iteration number.

According to \textit{Theorem 1}, $\vec{h}_\lambda^*$ gets infinitely close to the minimum of \eqref{eq3.2}, i.e., $\vec{h}^*$, as $\lambda$ goes to zero. This motivates us to adaptively decrease the smooth parameter during each call of \textit{Algorithm} \ref{alg.ogm}. The total procedure is summarized in \textit{Algorithm} \ref{alg.nesterov} which sets the smoothing parameter inversely proportional to the iteration number and thus improves the approximation iteratively. In \textit{Algorithm} \ref{alg.nesterov}, the current solution, i.e., $H_t$ and $W_t$, is used as a warm start to accelerate the convergence of \textit{Algorithm} \ref{alg.ogm} (see sentence 3 and 4).

\begin{algorithm}
\caption{Smoothing Method for MahNMF}
\label{alg.nesterov}
\begin{algorithmic}
\STATE \textbf{Input}: $X\in \mathrm{R}_+^{m\times n}$, $\lambda$, $\xi$, $\lambda_0$.
\STATE \textbf{Output}: $W_*$, $H_*$.
\STATE 1: Initialize $W_0\ge 0$, $H_0\ge 0$.
\STATE 2: For $k=0,1,2,...$
\STATE 3: \quad Solve $H_{t+1}$ by \textit{Algorithm} \ref{alg.ogm} with input $(X, W_t, H_t, ¦Ë_t, \epsilon_t^H)$.
\STATE 4: \quad Solve $W_{t+1}$ by \textit{Algorithm} \ref{alg.ogm} with input $(X, H_{t+1}, W_t, ¦Ë_t, \epsilon_t^W)$.
\STATE 5: \quad Update $\lambda_t=\frac{\lambda_0}{t+1}$.
\STATE 6: \quad Check the stopping condition \eqref{eq2.3}.
\STATE 7: End For
\STATE 8: $W_*=W_{t+1}$, $H_*=H_{t+1}$.
\end{algorithmic}
\end{algorithm}

\textit{Algorithm} \ref{alg.nesterov} recursively minimizes two smoothed objective functions, i.e., $f_{\lambda_t} (W_t,H)$ and $f_{\lambda_t} (W,H_{t+1})$, as $t$ goes to infinity. Although the generated point $\{(W_{t+1},H_{t+1})\}$ at the $t$-th iteration is not the minimum of the original sub-problems, the following \textit{Theorem 2} shows that $\{(W_{t+1},H_{t+1})\}$ do decrease the objective function $f(W,H)$ as $t$ goes to infinity. Since the objective function $f(W,H)$ is lower bounded, \textit{Algorithm} \ref{alg.nesterov} converges to an approximate solution of \eqref{eq1.1}.

\noindent
{\bf Theorem 2} {The sequence $(W_{t+1},H_{t+1})$ generated by \textit{Algorithm} \ref{alg.nesterov} decreases the objective function of MahNMF, i.e., for any $t\ge 0$, $f(W_{t+1},H_{t+1})\le f(W_t,H_t)$}

\noindent
{\bf Proof}. Without loss of generality, we take the $t$-th iteration round for example and show that \textit{Algorithm} \ref{alg.nesterov} decreases the objective function. Given $W_t$, the sentence 3 of \textit{Algorithm} \ref{alg.nesterov} implies that $H_{t+1}=\arg\min_{H\ge 0} f_(\lambda_t)(W_t,H)$. Therefore, we have $f_{\lambda_t} (W_t,H_{t+1})\le f_{\lambda_t} (W_t,H_t)$. According to \textit{Theorem 1}, we have
\begin{equation*}
    f(W_t,H_{t+1})=\sum_{j=1}^n f(W_t,H_{t+1}^j)\le \sum_{j=1}^n f_{\lambda_t}(W_t,H_{t+1}^j)+\frac{nD}{2}\lambda_t=f_{\lambda_t}(W_t,H_{t+1})+\frac{nD}{2}\lambda_t.
\end{equation*}
and $f_{\lambda_t}(W_t,H_t)\le f(W_t,H_t)$. Then we immediately have the following inequalities
\begin{equation}
    f(W_t,H_t)\ge f_{\lambda_t}(W_t,H_t)\ge f_{\lambda_t}(W_t,H_{t+1})\ge f(W_t,H_{t+1})-\frac{nD}{2}\lambda_t. \label{eq3.16}
\end{equation}
From \eqref{eq3.16}, we get that $f(W_t,H_t)-f(W_t,H_{t+1})+\frac{nD}{2}\lambda_t\ge f_{\lambda_t}(W_t,H_t)-f_{\lambda_t}(W_t,H_{t+1})$. By setting the precision for \textit{Algorithm} \ref{alg.ogm} as $\epsilon_t^H\le \frac{nD\lambda_t}{2}$ and using the objective-based stopping condition \eqref{eq3.11}, we have $f_{\lambda_t}(W_t,H_t)-f_{\lambda_t}(W_t,H_{t+1})\ge \frac{nD}{2}\lambda_t$, and thus $f(W_t,H_t)\ge f(W_t,H_{t+1})$. In a similar way, we can prove that $f(W_t,H_{t+1})\ge f(W_{t+1},H_{t+1})$. Therefore, $f(W_t,H_t)\ge f(W_{t+1},H_{t+1})$. This completes the proof. \hfill\BlackBox

In the proof of \textit{Theorem 2}, we need to set the precision of \textit{Algorithm} \ref{alg.ogm} as $\epsilon_t^H\le \frac{nD\lambda_t}{2}$ at the $t$-th iteration round. By substituting $\lambda_t=\frac{\lambda_0}{t+1}$, we have $\epsilon_t^H\le \frac{nD\lambda_0}{2(t+1)}$. Therefore, $\epsilon_t^H$ may go to zero as $t$ goes to infinity and this setting will make \textit{Algorithm} \ref{alg.ogm} fly without stopping. Fortunately, since \textit{Algorithm} \ref{alg.ogm} converges at the rate of $O(\frac{1}{k^2}$, it needs only $O(\sqrt{t})$ iterations to reach precision $\epsilon_t^H$ which is quite cheap. For example, suppose \textit{Algorithm} \ref{alg.nesterov} converges within $T\le 10^4$ iterations in its worst case, \textit{Algorithm} \ref{alg.ogm} needs only around $100$ iterations to reach precision $\epsilon_t^H$ when $t\le T$. This means that such an assumption is usually satisfied. Therefore, \textit{Algorithm} \ref{alg.nesterov} obtains an approximate solution for MahNMF when it stops.

The main time cost of \textit{Algorithm} \ref{alg.nesterov} is spent on sentences 3 and 4 which call \textit{Algorithm} \ref{alg.ogm} to successively update $H$ and $W$, respectively. Since the time complexity of \textit{Algorithm} \ref{alg.ogm} is $O(2mn(r+1))\times K$ and the iteration number $K$ depends on the specified precision, the time cost of the $t$-th iteration of \textit{Algorithm} \ref{alg.nesterov} is $O(2mn(r+1)\sqrt{t})$. Therefore, the total time complexity of \textit{Algorithm} \ref{alg.nesterov} is $O(2mn(r+1)\sum_{i=1}^T\sqrt{i})$, wherein $T$ is the iteration number. Empirically, $T$ is small, e.g., $T\le 100$, and thus \textit{Algorithm} \ref{alg.nesterov} converges fast.

In summary, the proposed Nesterov smoothing method-based algorithm costs less CPU time in each iteration than the proposed RRI method whose time complexity is $O(mnrlogm+mnr)\times K$, wherein $K$ are the iteration numbers of \textit{Algorithm} \ref{alg.rri}, but the RRI method converges in fewer iteration rounds because it finds a closed form solution for each variable of both factor matrices. Therefore, the performance of the proposed RRI algorithm and the smoothing-based algorithm is comparable for optimizing small scale MahNMF. However, the smoothing method is much more scalable than RRI due to its lower time complexity. Thus we suggest choosing the Nesterov smoothing method-based algorithm to optimizing MahNMF and its variants.

\section{MahNMF Extensions}
MahNMF provides a flexible framework for developing various algorithms for practical applications. In this section, we extend MahNMF by integrating box-constraint, manifold regularization, and group sparsity, and develop elastic net inducing MahNMF and symmetric MahNMF for several computer vision tasks.

\subsection{Box-Constrained MahNMF}
When the observations satisfy a box constraint such as $0\le \vec{x}_{(i)}\le 1$ for any $1\le i\le m$, it is reasonable to assume that the entries of $W$ and $H$ fall into the domain $[0,1]$. Based on this observation, we extend MahNMF to a box-constrained MahNMF (MahNMF-\textit{BC}) as follows
\begin{equation}
    \min_{0\le W_{(i,j)}\le 1,0\le H_{(i,j)}\le 1} f^{(bc)}(W,H)=\|X-W^T H\|_{l_1}. \label{eq4.1}
\end{equation}

It is natural to adopt the proposed Nesterov smoothing method to optimize \eqref{eq4.1}. It is surprising that \textit{Algorithm} \ref{alg.ogm} becomes much more efficient in this case. In particular, we replace the feasible set of the box-constrained non-negative least absolute deviation (NLAD-\textit{BC}) problem with $Q_1^{(bc)}=\{\vec{h}: 0\le \vec{h}_{(j)}\le 1\}$ and keep all the other definitions consistent. Instead of \eqref{eq3.8.1} and \eqref{eq3.8.2}, we solve the following two problems for the two auxiliary sequences:
\begin{equation}
    \vec{y}_k^{(bc)}=T_{Q_1^{(bc)}}(\vec{h}_k)={\arg\min}_{\vec{y}\in Q_1^{(bc)}}\{ <\nabla f_\lambda (\vec{h}_k),\vec{y}-\vec{h}_k>_1+\frac{L_\lambda}{2}\|\vec{y}-\vec{h}_k\|_1^2\}, \label{eq4.2.1}
\end{equation}
and
\begin{equation}
    \vec{z}_k^{(bc)}={\arg\min}_{\vec{z}\in Q_1^{(bc)}} \{\frac{L_\lambda}{\delta_1}d_1(\vec{z})+\sum_{i=0}^k \frac{i+1}{2} [f_\lambda(\vec{h}_i)+<f_\lambda(\vec{h}_i),\vec{z}-\vec{h}_i>_1]\}, \label{eq4.2.2}
\end{equation}
whose solutions are as follows:
\begin{equation}
    \vec{y}_k^{(bc)}=med (0,\vec{1},\vec{h}_k-\frac{1}{L_\lambda}\nabla f_\lambda(\vec{h}_k)), \label{eq4.3.1}
\end{equation}
and
\begin{equation}
    \vec{z}_k^{(bc)}=med (0,\vec{1},-\frac{1}{L_\lambda} \sum_{i=0}^k \frac{i+1}{2} \nabla f_\lambda (\vec{h}_i)). \label{eq4.3.2}
\end{equation}
By using the box constraint, it is quite easy to compute the bound of the prox-function for $Q_1^{(bc)}$, i.e., $D_1^{(bc)}=\frac{r}{2}$. Based on the obtained bound, it is easy to compute the dual function, i.e.,
\begin{equation*}
    \phi^{(bc)}(\vec{\mu})=\min_{\vec{h}} \{<W^T \vec{h}-\vec{x},\vec{\mu}>_1: \vec{h}\in Q_1^{(bc)}\}=<W^T \varphi_1(W\vec{\mu})-\vec{x},\vec{\mu}>_1.
\end{equation*}
Thanks to the closed-form dual function $\phi^{(bc)}(\vec{\mu})$, eq. \eqref{eq3.14} can be used to check the convergence of \textit{Algorithm} \ref{alg.ogm}. It greatly cuts down the time cost of \textit{Algorithm} \ref{alg.ogm} because the calculation of objective function $f_\lambda (W,\vec{h})$ is withdrawn.

The RRI method can also be naturally adopted to optimize MahNMF-\textit{BC} because the only difference between MahNMF and MahNMF-\textit{BC} is on their feasible sets. In particular, we keep all the other parts of \textit{Algorithm} \ref{alg.rri} consistent except sentence 8. After obtaining the piecewise point $p_{s^{c_j}}$, the closed form solution for sentence 8 for each variable in MahNMF-\textit{BC} is replaced by ${H_{k+1}}_{(l,j)}=med \{0,1,p_{s^{c_j}}\}$ for any $l\in \{1,...,r\}$ and $j\in \{1,...,n\}$.

\subsection{Manifold Regularized MahNMF}
When the observations distributed on the surface of a manifold are embedded in a high-dimensional space, one is interested in preserving the geometry structure in the learned low-dimensional space. Manifold regularization \citep{Tenenbaum2000} aims to preserve this geometry structure and constructs an adjacent graph $G$ to capture the neighbor relationship between one observation and a few of its nearest neighbors. By minimizing the distances between each observation and its corresponding nearest neighbors in the low-dimensional space, it preserves the geometry structure, i.e.,
\begin{equation}
    \min_H tr(HL^{(G)}H^T), \label{eq4.6}
\end{equation}
where $L^{(G)}$ is the Laplacian matrix of $G$. By combining \eqref{eq4.6} and \eqref{eq1.1}, we extend MahNMF to a manifold regularized MahNMF (MahNMF-\textit{M}), i.e.,
\begin{equation}
    \min_{W\ge 0,H\ge 0} f^{(M)}(W,H)=\|X-W^TH\|_M+\frac{\beta}{2}tr(HL^{(G)}H^T), \label{eq4.7}
\end{equation}
where $\beta>0$ is the tradeoff parameter.

MahNMF-\textit{M} can be solved by using alternating optimization over $W$ and $H$ with \textit{Algorithm} \ref{alg.ogm} and slightly modified \textit{Algorithm} \ref{alg.ogm}, respectively. We term the optimization procedure of $H$ a manifold regularized NLAD (NLAD-\textit{M}) problem. According to \citep{Guan2012}, the second term of $f^{(M)}(W,H)$ is convex and its gradient is Lipschitz continuous with constant $L^{(G)}$. Therefore, to solve NLAD-\textit{M}, sentences 5 and 6 in \textit{Algorithm} \ref{alg.ogm} should be replaced by
\begin{equation}
    Y_k^{(M)}=\max (0,H_k-\frac{1}{L_\lambda^{(m)}}\nabla f_\lambda^{(M)}(H_k)), \label{eq4.8.1}
\end{equation}
and
\begin{equation}
    Z_k^{(M)}=\max (0,-\frac{1}{L_\lambda^{(M)}} \sum_{i=0}^k \frac{i+1}{2} \nabla f_\lambda^{(M)}(H_k)), \label{eq4.8.2}
\end{equation}
where $\nabla f_\lambda^{(M)}(H_k)=\nabla f_\lambda(H_k)+\beta H_k L^{(G)}$ and $L_\lambda^{(M)}=L_\lambda+\beta L^{(G)}$.

In addition, the proposed RRI method can also be adopted to optimize MahNMF-\textit{M}. By using the residual matrix $Z$ defined in \eqref{eq2.4} and considering the $l$-th row of $H$, the objective function \eqref{eq4.7} can be equivalently rewritten as
\begin{equation}
    \min_{H_{(l)}\ge 0}f^{(M)}(W,H_{(l)})=\|Z-W_{(l)}^T H_{(l)}\|_M+\frac{\beta}{2} H_{(l)}L^{(G)}H_{(l)}^T. \label{eq4.8.5}
\end{equation}
Given all the variables in $H_{(l)}$ except $H_{(l,j)}$, eq. \eqref{eq4.8.5} is equivalent to
\begin{equation}
    \min_{H_{(l,j)}\ge 0}f^{(M)}(W,H_{(l,j)})=\sum_{i=1}^m |Z_{(i,j)}-W_{(l,i)}H_{(l,j)}|+\frac{\beta}{2}\sum_{a=1,a\ne j}^n S_{aj} (H_{(l,a)}-H_{(l,j)})^2, \label{eq4.8.6}
\end{equation}
where $S_{aj}>0$ is the $(a,j)$-th element of the similarity matrix for adjacent graph $G$. Since $f^{(M)}(W,H_{(l,j)})$ is actually a continuous, convex, and piecewise function, we can easily obtain its closed form solution based on the following \textit{Theorem 3}. Supposing the minimum of $f^{(M)}(W,H_{(l,j)})$ is $H_{(l,j)}^{'}$, the optimal solution of \eqref{eq4.8.6} is $H_{(l,j)}^*=\max (0,H_{(l,j)}^{'})$. Note that $H_{(l,j)}^{'}$ is selected from the piecewise point set $\{\frac{Z_{(i,j)}}{W_{(l,i)}}©¦i=1,...,m\}$ according to \textit{Theorem 3}. If $Z$ contains all zero, then the optimal solutions of \eqref{eq4.8.5} will be trivial, i.e., $H_{(l)}^*=0$. To overcome this problem, we need to initialize both $W$ and $H$ by a small value, e.g., $10^{-10}$. In our experiment, this initialization strategy works well.

\noindent
{\bf Theorem 3} {Given $f(x)=\sum_{i=1}^m |a_i (x-x_i)|+b(x-d)^2$, wherein $a_i>0$ and $b>0$ and $x_1<\cdots<x_m$. Define $k_{i+1}=k_i+2a_i$ and $k_1=-a_1-\cdots-a_m$. If $2b(x_i-d)+k_{i+1}\le 0$ and $2b(x_{i+1}-d)+k_{i+1}>0$ for a some $i$, then the minimum of $f(x)$ is
\[
    x_*=\left \{\begin{array}{c@{\;}l} d-\frac{k_1}{2b},&i=0 \\
    d-\frac{k_{m+1}}{2b},&i=m \\
    \max\{x_i,d-\frac{k_{i+1}}{2b}\}, &i\in \{1,...,m-1\}.
    \end{array} \right.
\]
}

\noindent
{\bf Proof}. Using algebra, $f(x)$ can be written as a piecewise quadratic function as follows:
\begin{equation}
    f(x)=\left \{\begin{array}{c@{\;}l}
    b(x-d)^2+k_1 x+c_1, &x_0< x\le x_1 \\
    b(x-d)^2+k_2 x+c_2, &x_1\le x\le x_2 \\
    \vdots \\
    b(x-d)^2+k_m x+c_m, &x_{m-1}\le x\le x_m \\
    b(x-d)^2+k_{m+1} x+c_{m+1}, &x_m\le x< x_{m+1} \end{array}\right., \label{eq4.8.3}
\end{equation}
where $c_1=a_1 x_1+\cdots+a_m x_m$ and $c_{i+1}=c_i-2a_i x_i$. Here we define $x_0=-\infty$ and $x_{m+1}=\infty$ for the simplicity of presentation. Since the first part of $f(x)$ is convex and the second part is strongly convex, $f(x)$ is totally strongly convex, and thus it has an unique minimum $x_*$. According to \eqref{eq4.8.3}, we obtain the slope of $f(x)$ as follows:
\begin{equation}
    f^{'}(x)=\left\{\begin{array}{c@{\;}l}
    2b(x-d)+k_1, &x_0<x\le x_1 \\
    2b(x-d)+k_2, &x_1\le x\le x_2 \\
    \vdots \\
    2b(x-d)+k_m, &x_{m-1}\le x\le x_m \\
    2b(x-d)+k_{m+1}, &x_m\le x<x_{m+1}\end{array}\right.. \label{eq4.8.4}
\end{equation}
Here we define $f^{'}(x_0)=-\infty$ and $f^{'}(x_{m+1})=\infty$ for the convenience of derivation. It is obvious that $f^{'}(x)$ is non-continuous; we define the left slope and right slope at each piecewise point $x_i$ as $f_-^{'}(x_i)=2b(x_i-d)+k_i$ with $i\in\{1,...,m+1\}$, and $f_+^{'}(x_i)=2b(x_i-d)+k_{i+1}$ with $i\in\{0,...,m\}$, respectively. Since $f(x)$ is continuous and strongly convex, $x_*$ is unique and it appears at the point that first changes the sign of $f^{'}(x)$. Suppose $f_+^{'}(x_i)\le 0$ and $f_+^{'}(x_{i+1})\ge 0$, wherein $i\in {0,...,m}$, we have $x_i\le x_*\le x_{i+1}$. If $i=0$, we have $x_*=d-\frac{k_1}{2b}$ because $f(x)$ is a quadratic function on the set $(x_i,x_{i+1}]$. If $i=m$, we have $x_*=d-\frac{k_{m+1}}{2b}$ because $f(x)$ is a quadratic function on the set $[x_i,x_{i+1})$. If $i\in\{1,...,m-1\}$, $f(x)$ is a quadratic function on the set $[x_i,x_{i+1}]$, we have $x_*=med\{x_i,x_{i+1},d-\frac{k_{m+1}}{2b}\}$. Since $f_+^{'}(x_{i+1})\ge 0$, we have $d-\frac{k_{m+1}}{2b}<x_{i+1}$, then $x_*=\max\{x_i,d-\frac{k_{i+1}}{2b}\}$. It completes the proof. \hfill\BlackBox

Although RRI can be applied to the optimization of MahNMF-\textit{M}, it is time-consuming because the variables must be updated one by one. We suggest the proposed Nesterov smoothing method for optimizing MahNMF-\textit{M}.

\subsection{Group Sparse MahNMF}
Since NMF does not explicitly guarantee sparse representation, Hoyer proposed sparseness-constrained NMF (NMFsc, \citep{Hoyer2004}) to incorporate sparseness constraint on single or both factor matrices. Recent results show that many data sets are inherently structured as groups \citep{Bengio2009}\citep{Huang2009}, i.e., some of the data items or features that belong to the same group share the same sparsity pattern. For example, different types of features such as pixels, gradient-based features, and color-based features of an image can be considered as different groups. Such prior knowledge of group sparsity greatly improves the effectiveness of sparse representation and has be successfully applied in many methods, e.g., group Lasso \citep{Yuan2006}. It motivates us to introduce group sparsity to explicitly improve the sparse representation of MahNMF. The objective of group sparse MahNMF (MahNMF-\textit{GS}) is as follows:
\begin{equation}
    \min_{W\ge 0,H\ge 0} \|X-W^T H\|_M,s.t.,\forall\rho\in G_W,\|{W^{[\rho]}}^T\|_{1,p}\le \gamma_W,\forall\rho\in G_H,\|{H^{[\rho]}}^T\|_{1,p}\le\gamma_H, \label{eq4.9}
\end{equation}
or
\begin{equation}
    \min_{W\ge 0,H\ge 0} \|X-W^T H\|_M+\eta_W \sum_{\rho\in G_W} \|{W^{[\rho]}}^T\|_{1,p}+\eta_H \sum_{\rho\in G_H}\|{H^{[\rho]}}^T\|_{1,p}, \label{eq4.10}
\end{equation}
where $X^{[\rho]}$ signifies the columns of $X$ indexed by group $\rho$, and $G_W\subset2^{\{1,...,m\}}$ and $G_H\subset2^{\{1,...,n\}}$ are the grouping sets of columns of $W$ and $H$, and $\gamma_W$ and $\gamma_H$ control the group sparsity of $W$ and $H$, respectively. The tradeoff parameters $\eta_W>0$ and $\eta_H>0$ control the group sparsity over $W$ and $H$, respectively. The group sparsity is usually defined by using $L_{1,p}$-norm which is defined as $\|X\|_{1,p}=\sum_{j=1}^b|\|H^{[\{j\}]}\|_p|$  for any $X\in\mathrm{R}^{a\times b}$, wherein $p\ge 1$. Usually, we choose $p=2,\infty$ for group sparsity. In the following section, we will show that both \eqref{eq4.9} and \eqref{eq4.10} can be solved by slightly modifying the proposed Nesterov's smoothing method.

To solve \eqref{eq4.9}, we modified \textit{Algorithm} \ref{alg.nesterov} by redefining the feasible set of $W$ and $H$ as
\begin{equation}
    Q_W^{(gs)}=\{W\in \mathrm{R}_+^{r\times m}©¦\forall\rho\in G_W,\|{W^{[\rho]}}^T\|_{1,p}\le \gamma_W,G_W\subset2^{\{1,...,m\}}
\end{equation}
and
\begin{equation}
    Q_H^{(gs)}=\{H\in \mathrm{R}_+^{r\times n}©¦\forall\rho\in G_H,\|{H^{[\rho]}}^T\|_{1,p}\le \gamma_H,G_H\subset2^{\{1,...,n\}}.
\end{equation}
Based on the alternating optimization method, given $W$, $H$ can be optimized by cycling on variables indexed by $G_H$ because the groups are  non-overlapping. For any group $\rho\in G_H$, the objective for optimizing $H^{[\rho]}$ is
\begin{equation}
    \min_{H^{[\rho]}\in Q_{H^{[\rho]}}^{(gs)}} \|X^{[\rho]}-W^T H^{[\rho]}\|_M, \label{eq4.11}
\end{equation}
where $Q_{H^{[\rho]}}^{(gs)}=\{H^{[\rho]}\in\mathrm{R}_+^{r\times n_\rho}|\|{H^{[\rho]}}^T\|_{1,p}\le \gamma_H\}$. It is obvious that $Q_{H^{[\rho]}}^{(gs)}$ is closed and convex set, and thus \eqref{eq4.11} can be solved by slightly modifying \textit{Algorithm} \ref{alg.ogm}. Particularly, at the $k$-th iteration, the sequences $Y_k$ and $Z_k$ can be obtained by solving the following problems:
\begin{equation}
    Y_k^{[\rho]}={\arg\min}_{Y^{[\rho]}\in Q_{H^{[\rho]}}^{(gs)}} \{<f_\lambda(W,H_k^{[\rho]}),Y^{[\rho]}-H_k^{[\rho]}>_1+\frac{L_\lambda}{2}\|Y^{[\rho]}-H_k^{[\rho]}\|_F^2\}, \label{eq4.12.1}
\end{equation}
and
\begin{align}
    Z_k^{[\rho]}={\arg\min}_{Z^{[\rho]}\in Q_(H^{[\rho]})^{(gs)}}\{\frac{L_\lambda}{2\delta_1} \|Z^{[\rho]}\|_F^2&+\sum_{i=0}^k \frac{i+1}{2}[f_\lambda(W,H_i^{[\rho]}) \nonumber \\
    &+<\nabla f_\lambda(W,H_i^{[\rho]}),Z^{[\rho]}-H_i^{[\rho]}>_1]\}, \label{eq4.12.2}
\end{align}
respectively. Both \eqref{eq4.12.1} and \eqref{eq4.12.2} essentially minimize a quadratic function over a convex set, and thus they can be solved by projecting the minimum of the corresponding quadratic functions as follows:
\begin{equation}
    Y_k^{[\rho]}=\prod_{Q_{H^{[\rho]}}^{(gs)}} (H_k^{[\rho]} -\frac{1}{L_\lambda}\nabla f_\lambda (W,H_k^{[\rho]})), \label{eq4.13.1}
\end{equation}
and
\begin{equation}
    Z_k^{[\rho]}=\prod_{Q_{H^{[\rho]}}^{(gs)}} (-\frac{1}{L_\lambda} \sum_{i=0}^k \frac{i+1}{2} \nabla f_\lambda (W,H_i^{[\rho]})), \label{eq4.13.2}
\end{equation}
where $\prod_{Q_{H^{[\rho]}}^{(gs)}}(X)$ projects $X$ onto $Q_{H^{[\rho]}}^{(gs)}$. The projection operator can be defined as
\begin{equation}
    \min_{X\ge 0,\|X\|_{1,p}\le\gamma_H} \frac{1}{2} \|X-\tilde{X}\|_F^2. \label{eq4.14}
\end{equation}
According to \citep{Tandon2010}, the non-zero entries in the optimal solution, namely $X_*$, of (4.14) share the same signs as those in $\tilde{X}$. Therefore, eq. \eqref{eq4.14} can be solved by projecting the absolute of $\tilde{X}$ onto the $l_{1,p}$ ball, i.e.,
\begin{equation}
    X_*=\emph{proj}_{\gamma_H}^p (\max(0,\tilde{X})). \label{eq4.15}
\end{equation}
When $p=2$, the projection can be done by using Berg's algorithm \citep{Berg2008} in $O(rn_\rho)$ time. When $p=\infty$,  the projection is completed by using Quattoni's algorithm \citep{Quattoni2009} in $O(rn_\rho \log n_\rho)$ time. Therefore, the proposed Nesterov smoothing method-based algorithm can be applied to optimizing \eqref{eq4.9} without increasing the time complexity. Moreover, the following section will show that it can also be adopted to optimizing \eqref{eq4.10}.

Although the objective function of \eqref{eq4.10} is non-convex with respect to $W$ and $H$ simultaneously, it is convex with respect to either $W$ or $H$. Therefore, eq. \eqref{eq4.10} can be solved by alternatively optimizing $W$ and $H$. Take the sub-problem of optimizing $H$ (called group sparse NLAD or NLAD-\textit{GS} for short) for example, its objective function is as follows:
\begin{equation}
    \min_{H\ge 0} \|X-W^T H\|_M+\eta_H \sum_{\rho\in G_H} \|{H^{[\rho]}}^T\|_{1,p}. \label{eq4.16}
\end{equation}
Since both $\|X-W^T H\|_M$ and $\sum_{\rho\in G_H}\|{H^{[\rho]}}^T\|_{1,p}$ are convex, eq. \eqref{eq4.16} has an optimal solution. However, it is involved because neither $\|X-W^T H\|_M$ nor $\sum_{\rho\in G_H}\|{H^{[\rho]}}^T\|_{1,p}$ is smooth. Fortunately, the OGM method (see \textit{Algorithm} \ref{alg.ogm}) can be slightly modified to solve it efficiently. By using the smoothing function $f_\lambda(W,H)$, eq. \eqref{eq4.16} can be approximated by
\begin{equation}
    \min_{H\ge 0} f_\lambda(W,H)+\eta_H\sum_{\rho\in G_H}\|{H^{[\rho]}}^T\|_{1,p}. \label{eq4.17}
\end{equation}
Since $G_H$ is non-overlapping, $f_\lambda(W,H)$ is separable, i.e., $f_\lambda(W,H)=\sum_{\rho\in G_H} f_\lambda(W,H^{[\rho]})$, Eq. \eqref{eq4.17} can be solved by recursively optimizing each group of variables, i.e.,
\begin{equation}
    \min_{H^{[\rho]}\ge 0} f_\lambda(W,H^{[\rho]})+\eta_H\|{H^{[\rho]}}^T\|_{1,p}, \label{eq4.18}
\end{equation}
where $\rho\in G_H$. In order to solve \eqref{eq4.18} by using the OGM method, we construct additional two auxiliary sequences, i.e., $Y_k^{[\rho]}$ and $Z_k^{[\rho]}$, wherein $k\ge 0$ is the iteration counter. Since OGM essentially constructs the 'Y' sequence by optimizing a linear approximation of $f_\lambda(W,\cdot)$ at $H_k^{[\rho]}$ regularized by a quadratic proximal term, we propose to construct $Y_k^{[\rho]}$ by optimizing the following objective function
\begin{equation}
    Y_k^{[\rho]}={\arg\min}_{Y^{[\rho]}\in Q_1} \{<\nabla f_\lambda (W,H_k^{[\rho]}),Y^{[\rho]}-H_k^{[\rho]}>_1+\frac{L_\lambda}{2}\|Y^{[\rho]} -H_k^{[\rho]}\|_1^2+\eta_H\|{Y^{[\rho]}}^T\|_{1,p}]\}. \label{eq4.19}
\end{equation}
Because \eqref{eq4.19} employs no approximation on the non-smooth part, such approximation will not decrease the convergence rate of \textit{Algorithm} \ref{alg.ogm} if \eqref{eq4.19} can be efficiently solved. Fortunately, the answer is positive. Using algebra, eq. \eqref{eq4.19} can be equivalently rewritten as
\begin{align}
    Y_k^{[\rho]}&={\arg\min}_{Y^{[\rho]}\in Q_1} \{\frac{L_\lambda}{2}\|Y^{[\rho]}-(H_k^{[\rho]}-\frac{1}{L_\lambda}\nabla f_\lambda(W,H_k^{[\rho]}))\|_1^2 \nonumber \\
    &-\frac{1}{2L_\lambda}\|\nabla f_\lambda(W,H_k^{[\rho]})\|_1^2+\eta_H\|{Y^{[\rho]}}^T\|_{1,p}\} \nonumber \\
    &={\arg\min}_{Y^{[\rho]}\in Q_1} \{\frac{1}{2}\|Y^{[\rho]}-(H_k^{[\rho]}-\frac{1}{L_\lambda}\nabla f_\lambda(W,H_k^{[\rho]}))\|_1^2+\frac{\eta_H}{L_\lambda}\|{Y^{[\rho]}}^T\|_{1,p}\}. \label{eq4.20}
\end{align}

According to \citep{Tandon2010}, eq. \eqref{eq4.20} reduces to the well-known proximity operator problem, i.e.,
\begin{equation}
    Y_k^{[\rho]}={\arg\min}_{Y^{[\rho]}}\{\frac{1}{2}\|Y^{[\rho]}-\prod_{Q_1}(H_k^{[\rho]}-\frac{1}{L_\lambda}\nabla f_\lambda(W,H_k^{[\rho]}))\|_1^2+\frac{\eta_H}{L_\lambda}\|{Y^{[\rho]}}^T\|_{1,p}\}. \label{eq4.21}
\end{equation}
This problem can be solved by first solving its dual problem and projecting the solution back to solve the primal problem. When $p=2$, the dual problem of \eqref{eq4.21} can be easily solved by normalization. When $p=\infty$, its dual problem is equivalent to $l_1$-norm projection that can be efficiently solved by using Duchi's method in linear time \citep{Duchi2008}.

Since OGM constructs the 'Z' sequence by optimizing a combination of the linear approximations of $f_\lambda (W,\cdot)$ at the historical search points regularized by a quadratic term, similar to \eqref{eq4.19}, we can construct $Z_k^{[\rho]}$ by adding the non-smooth part to each linear approximation, i.e.,
\begin{align}
    Z_k^{[\rho]} &={\arg\min}_{Z^{[\rho]}}\{\frac{L_\lambda}{2}\|Z^{[\rho]}\|_F^2+\sum_{i=0}^k\frac{i+1}{2}(f_\lambda(W,H_i^{[\rho]})+<\nabla f_\lambda(W,H_i^{[\rho]}),Z^{[\rho]}-H_i^{[\rho]}>_1 \nonumber \\
    &+\eta_H\|{Z^{[\rho]}}^T\|_{1,p})\} \nonumber \\
    &= {\arg\min}_{Z^{[\rho]}}\{\frac{L_\lambda}{2}\|Z^{[\rho]}\|_F^2+\sum_{i=0}^k\frac{i+1}{2}(f_\lambda(W,H_i^{[\rho]})+<\nabla f_\lambda(W,H_i^{[\rho]}),Z^{[\rho]}-H_i^{[\rho]}>_1) \nonumber \\
    &+\frac{\eta_H(k+1)(k+2)}{4}\|{Z^{[\rho]}}^T\|_{1,p}\} \nonumber \\
    &= {\arg\min}_{Z^{[\rho]}}\{\frac{1}{2}\|Z^{[\rho]}-\prod_{Q_1}(-\frac{1}{L_\lambda} \sum_{i=0}^k \frac{i+1}{2}\nabla f_\lambda (W,H_i^{[\rho]}))\|_1^2 \nonumber \\
    &+\frac{\eta_H (k+1)(k+2)}{4L_\lambda}\|{Z^{[\rho]}}^T\|_{1,p}\}, \label{eq4.22}
\end{align}
which is also a proximity operator problem and can be efficiently solved in a similar way to \eqref{eq4.21}.

By replacing the sentences 5 and 6 in \textit{Algorithm} \ref{alg.ogm} with \eqref{eq4.21} and \eqref{eq4.22}, respectively, eq. \eqref{eq4.18} can be solved by modifying \textit{Algorithm} \ref{alg.ogm} and the new algorithm converges at the rate of $O(\frac{1}{k^2})$. We leave the proof to future work due to the limit of space. Empirical results show that the smoothing method for MahNMF-\textit{GS} converges rapidly. MahNMF-\textit{GS} is useful in many problems especially in multi-view learning. We will evaluate its effectiveness in the following section.

\subsection{Elastic Net Inducing MahNMF}
MahNMF decomposes a given non-negative matrix into a non-negative low-rank part and a sparse part. However, if there is a group of nonzero variables in the sparse part which are highly correlated, MahNMF tends to select only one variable from the group regardless which one is selected. That is because MahNMF introduces the sparity over the sparse part in a same way as Lasso \citep{Tibshirani1996}. In contrast, Zou and Hastie \citep{Zou2005} proposed an elastic net method to take into account the grouping effect of variables in regression. Elastic net minimizes the least squares loss function combined with both $l_1$ norm and $l_2$ norm over the coefficients and thus selects groups of correlated variables. Here, we introduce the main idea of Elastic net into MahNMF to take its advantage. In particular, we expect the highly correlated nonzero variables in the sparse part to be grouped by minimizing both the Manhattan distance and Euclidean distance between a non-negative matrix $X$ and its non-negative low-rank approximation $W^T H$ simultanously. We termed this extension elastic net inducing MahNMF (MahNMF-\textit{EN}) whose objective function is
\begin{equation}
    \min_{W\ge 0,H\ge 0} f^{(en)}(W,H)=\|X-W^T H\|_M+\frac{\alpha}{2} \|X-W^T H\|_F^2, \label{eq4.22.1}
\end{equation}
where $\alpha>0$ balances the Manhattan distance and the Euclidean distance between $X$ and $W^T H$.

Since $f^{(en)} (W,H)$ is composed of a smooth part and a non-smooth part, the Nesterov smoothing method can be naturally applied to optimizing \eqref{eq4.22.1}. Since $f^{(en)}(W,H)$ is non-convex, we solve \eqref{eq4.22.1} by recursively optimizing $W$ and $H$ until convergence. Given $W$, $H$ is updated by solving the elastic net inducing non-negative least absolute deviation (NLAD-\textit{EN}) problem and $W$ can be updated similarly with H fixed. According to \citep{Guan2012}, the second term $\frac{\alpha}{2} \|X-W^T H\|_F^2$ of \eqref{eq4.22.1} is convex and its gradient is Lipschitz continuous with constant $\alpha\|WW^T\|_2$, wherein $\|\cdot\|$ signifies the matrix spectral norm. Therefore, the proposed \textit{Algorithm} \ref{alg.ogm} can be applied to optimizing NLAD-\textit{EN} by replacing the gradient and Lipschitz constant with $\nabla f_\lambda^{(en)}(\vec{h}_k)=\nabla f_\lambda (\vec{h}_k)+\alpha W(W^T\vec{h}_k-\vec{x})$ and $L_\lambda^{(en)}=L_\lambda+\alpha \|WW^T\|_2$, respectively. That is to say, the proposed Nesterov smoothing method, i.e., \textit{Algorithm} \ref{alg.nesterov}, can be naturally adopted to solve MahNMF-\textit{EN} without increasing the time complexity.

In MahNMF-\textit{EN}, the trade-off parameter ¦Á plays a critical role to control the grouping effect of the sparse part. This parameter can be carefully selected based on the strategy introduced in \citep{Zou2005}.

\subsection{Symmetric MahNMF}
In spectral clustering, the data matrix $X$ is the Laplacian matrix or normalized Laplacian matrix of the specified adjacent graph. In these cases, the data matrix is symmetric and it is reasonable to cut down the number of variables by assuming $W=H$. Inspired by spectral clustering, we extend MahNMF to symmetric MahNMF (or MahNMF-\textit{SYM} for short), i.e.,
\begin{equation}
    \min_{H\ge 0}\|X-HH^T\|_M, \label{eq4.23}
\end{equation}
where $H\in \mathrm{R}_+^{n\times r}$ and $r\ll n$.

Since \eqref{eq4.23} is neither convex nor smooth, it is an involved problem. Fortunately, the proposed RRI method can be applied to successively update each variable of $H$ in a closed form solution. In particular, eq. \eqref{eq4.23} can be equivalently rewritten as $X\approx H^{(1)}{H^{(1)}}^T+\cdots+H^{(r)}{H^{(r)}}^T$. To optimize each column $H^{(c)}$ of $H$, wherein $c\in \{1,...,r\}$, we fix the other columns and solve the following problem
\begin{equation}
    \min_{H^{(c)}\ge 0} \|Z-H^{(c)}{H^{(c)}}^T\|_M, \label{eq4.23.1}
\end{equation}
where $Z=X-\sum_{i\ne c}^r H^{(i)}{H^{(i)}}^T$ is the residual matrix. Moreover, Eq. \eqref{eq4.23.1} can be solved by successively updating each variable. Considering variable $H_{(j,c)}$ with other variables fixed, we have
\begin{equation}
    \min_{H_{(j,c)}\ge 0} |Z_{(j,j)}-H_{(j,c)}^2|+\sum_{i\ne j}^n |Z_{(i,j)}-H_{(i,c)}H_{(j,c)}|. \label{eq4.24}
\end{equation}
Assuming $H_{(j,c)}^2\ge Z_{(j,j)}$, eq. \eqref{eq4.24} can be rewritten in a similar form to \eqref{eq4.8.6}, and thus it can be updated in a closed form solution by using \textit{Theorem 3}. The inequality $H_{(j,c)}^2\ge Z_{(j,j)}$ means that $\sum_i^r H_{(j,i)}^2\ge X_{(j,j)}$, which can be easily satisfied by normalizing $X$. RRI converges fast because the variables are updated in a closed form solution.

MahNMF-\textit{SYM} is useful in practice especially for spectral clustering. In the following section, we will show that MahNMF-\textit{SYM} can be successfully applied to image segmentation and discuss its relationship to normalized cut \citep{Shi2000}.

\begin{figure*}[ht]
\centering
\includegraphics[width=1.0\linewidth]{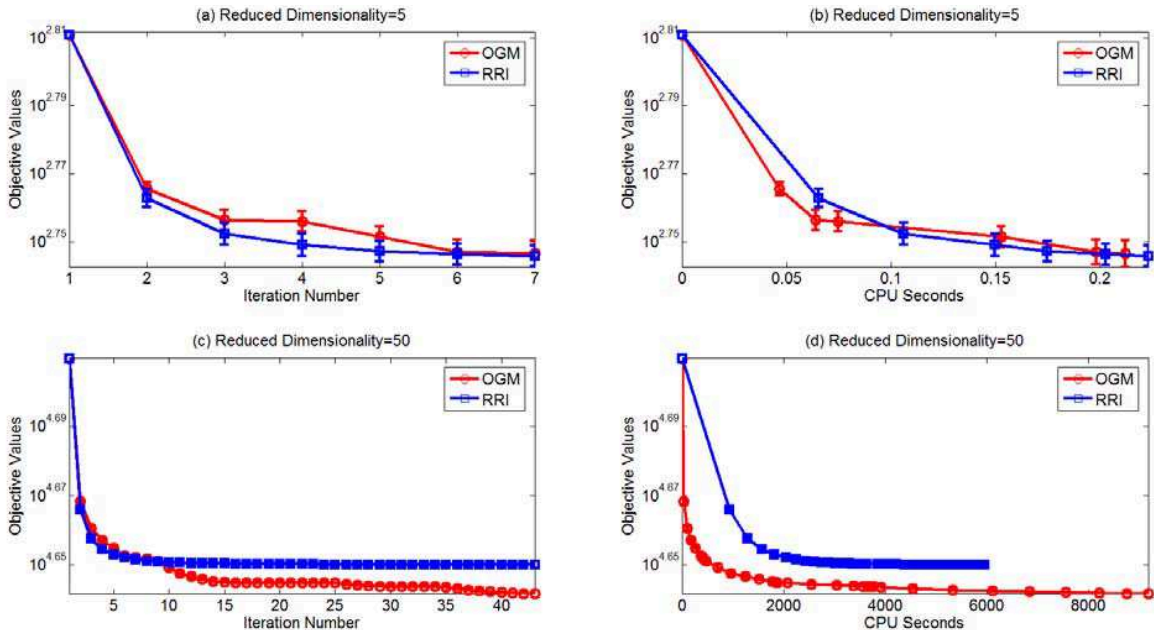}
\caption{Objective values versus iteration numbers and CPU seconds on $100\times 50$-D (a \& b) and $1000\times 500$-D (c \& d) synthetic datasets.}
\label{fig.eff-synthetic}
\end{figure*}

\section{Experimental Results}
In this section, we first compare the efficiency of the rank one residual iteration (RRI) method with that of the Nesterov smoothing method for optimizing MahNMF on both synthetic and real-world datasets. Subsequently, we study the effectiveness and robustness of MahNMF by comparing it with EucNMF and KLNMF by conducting face recognition and clustering on both Yale B and PIE datasets. We conduct image segmentation with MahNMF-SYM to study its clustering effectiveness. We then study the sparse and low-rank decomposition capability of MahNMF by conducting background and illumination modeling on video sequences and challenging face images dataset and comparing it with both robust principal component analysis (RPCA, \citep{Candes2011}) and GoDec, \citep{Zhou2011}. Finally, we apply the MahNMF-\textit{GS} algorithm to multi-view learning on two challenging datasets including VOC Pascal 07 and Mirflickr to show its effectiveness.

In this experiment, we use the multiplicative update rule \citep{Lee1999} to optimize KLNMF which stops when the indices of the column maximums of H do not change for $40$ consecutive iterations. We apply the efficient NMF solver NeNMF \citep{Guan2012} to optimize EucNMF and use the projected gradient norm-based criterion as a stopping condition with the precision setting to $10^{-8}$. The MahNMF stops until the stopping condition \eqref{eq2.3} is satisfied with precision setting to $0.1$. The smoothness parameter in \textit{Algorithm} \ref{alg.nesterov} is initialized to $\lambda_0=.1$ to guarantee fast convergence in the first steps and decrease dramatically to improve the approximation accuracy.

\begin{figure*}[ht]
\centering
\includegraphics[width=1.0\linewidth]{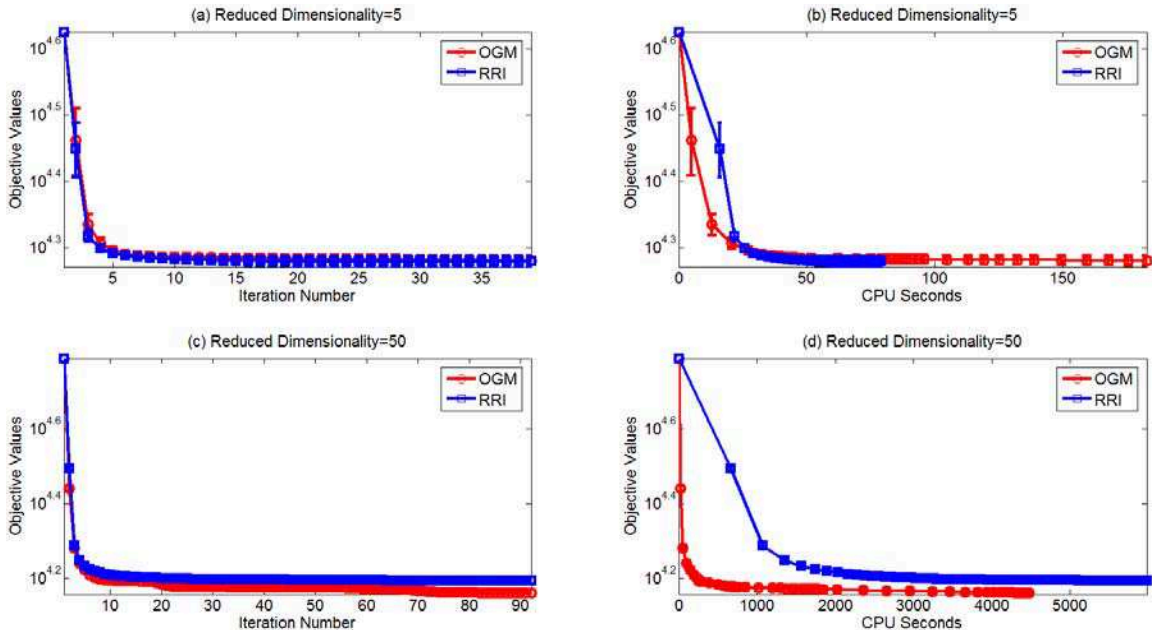}
\caption{Objective values versus iteration numbers and CPU seconds on the Yale B (a \& b) and PIE (c \& d) datasets.}
\label{fig.eff-real-world}
\end{figure*}

\subsection{RRI versus Nesterov's Smoothing Method}
As discussed above, both the rank-one residual iteration (RRI) method and Nesterov smoothing method, i.e., OGM, can be applied to the optimization of MahNMF and its extensions. One may expect to have to choose between them in practical applications. This section tries to address this issue by comparing their efficiency on both synthetic and real-world datasets.

To study the scalabilities of both algorithms, we conducted them on $100\times 50$-D and $1000\times 500$-D dense matrices, and set the reduced dimensionalities to $5$ and $50$, respectively. For fairness of comparison, both algorithms start from an identical randomly generated dense matrix. Since MahNMF is non-convex, the initial point has a high impact on the obtained solution. To filter this initialization impact, we repeat this experiment ten times and compare their average objective values and standard deviations versus iteration number and CPU seconds in Figure \ref{fig.eff-synthetic}.

Figure \ref{fig.eff-synthetic} (a) and (b) show that RRI converges in fewer iteration rounds and CPU seconds than OGM on the small scale data matrix. This is because a single iteration of RRI and OGM has a comparable time cost on a small scale data matrix while OGM obtains a closed form solution for each variable and further reduces the objective function. However, when the scale of the data matrix increases, OGM costs far fewer CPU seconds in each iteration round than RRI because it converges much more rapidly than RRI (see Figure \ref{fig.eff-synthetic} (c) and (d)). It confirms that OGM is more scalable than RRI.

We also conducted both RRI and OGM on two real-world datasets, i.e., Yale B \citep{Georghiades2001} and PIE \citep{Sim2003} face image datasets. The extended Yale B and PIE datasets contain $16,128$ and $41,368$ face images taken from $38$ and $68$ individuals, respectively. Each image is cropped to $32\times 32$ pixels and reshaped to a long vector. In this experiment, we randomly select seven images of each individual and construct a $1024\times 266$-dimensional matrix and a $1024\times 476$-dimensional data matrix, respectively, for MahNMF learning. Similar to the above experiment, we set the reduced dimensionality to $5$ and $50$, respectively. Figure \ref{fig.eff-real-world} gives their objective values and standard deviations versus iteration number and CPU seconds. From Figure \ref{fig.eff-real-world}, we have the same observations as those obtained from Figure \ref{fig.eff-synthetic}.

In summary, when the scale of the data matrix and reduced dimensionality are ordinarily small we suggest optimizing MahNMF by using RRI. When the scale of the data matrix and reduced dimensionality are relatively large we suggest optimizing MahNMF by using Nesterov's smoothing method to take its advantage of scalability.

\subsection{Face Recognition}
We study the data representation capacity of MahNMF by conducting face recognition experiments on two challenging face image datasets including Yale B \citep{Georghiades2001} and PIE \citep{Sim2003} and making a comparison with two traditional NMF algorithms including EucNMF and KLNMF. We randomly select seven images of each individual to construct the training set $X_{train}$ and the remaining images make up the test set $X_{test}$. To eliminate the effectiveness of random selection, we repeat this trial ten times and report the average accuracy and standard deviation. All the NMF algorithms are used to factorize the training set into the product of basis $W$ and the compact representation $H_{train}$, i.e., $X_{train}\approx W^T H_{train}$.

\begin{algorithm}
\caption{Traditional NMF-based face recognition}
\label{alg.trd-frc}
\begin{algorithmic}
\STATE 1: Compute $H_train={W^T}^{\dagger} X_{train}$, $H_{test}={W^T}^{\dagger}X_{test}$.
\STATE 2: For $i=1,2,...,n_{test}$
\STATE 3: \quad Compute $j={\arg\min}_l \{\|H_{train}^{(l)}-H_{test}^{(i)}\|_{l_2}\}$.
\STATE 4: \quad Transfer $X_{train}^{(j)}$'s label to $X_{test}^{(i)}$.
\STATE 5: End For
\end{algorithmic}
\end{algorithm}

The traditional NMF-based face recognition method obtains the representations of the test set by projecting $X_{test}$ onto the learned space as $H_test={W^T}^{\dagger} X_{test}$, wherein ${W^T}^{\dagger}$ is the pseudo inverse of $W^T$. \textit{Algorithm} \ref{alg.trd-frc} summarizes this method which transfers the label of Euclidean distance-based nearest neighbor (NN) in the training set to the given test sample. Although this method is efficient and easy to implement, the pseudo inverse operator may bring in negative elements and thus it is not robust to outliers.

To overcome the aforementioned drawback, Sandler and Lindenbaum \citep{Sandler2011} suggested another NMF-based face recognition method for the corrupted training and test set (see \textit{Algorithm} \ref{alg.san-frc}). This method finds the best compact representation of the test sample on the learned basis and transfers the label of cosine distance-based nearest neighbor in the training set to the given test sample. In \textit{Algorithm} \ref{alg.san-frc}, $D(\cdot,\cdot)$ is determined based on the NMF algorithm used, e.g. Manhattan distance for MahNMF. The accuracy is calculated as the percentage of test samples that are correctly classified.

\begin{algorithm}
\caption{Sandler-Lindenbaum's NMF-based face recognition}
\label{alg.san-frc}
\begin{algorithmic}
\STATE 1: Compute $H_test={\arg\min}_{H_{test}\ge 0} D(X_{test},W^TH_{test}$.
\STATE 2: For $i=1,2,...,n_{test}$
\STATE 3: \quad Compute $j={\arg\min}_l \{\frac{<H_{train}^{(l)},H_{test}^{(i)}>}{\|H_{train}\|_{l_2}\|H_{test}\|_{l_2}}\}$.
\STATE 4: \quad Transfer $X_{train}^{(j)}$'s label to $X_{test}^{(i)}$.
\STATE 5: End For
\end{algorithmic}
\end{algorithm}

To evaluate the robustness of MahNMF, we add five types of outliers including occlusion, Laplace noise, Salt \& Pepper noise, Gaussian noise and Poisson noise to the training set. In the classification stage, we conduct \textit{Algorithm} \ref{alg.trd-frc} and \textit{Algorithm} \ref{alg.san-frc} on the clean and contaminated test sets, respectively, to evaluate the robustness of MahNMF under different settings. The experimental results of MahNMF under both settings are encouraging.

\begin{figure*}[ht]
\centering
\includegraphics[width=1.0\linewidth]{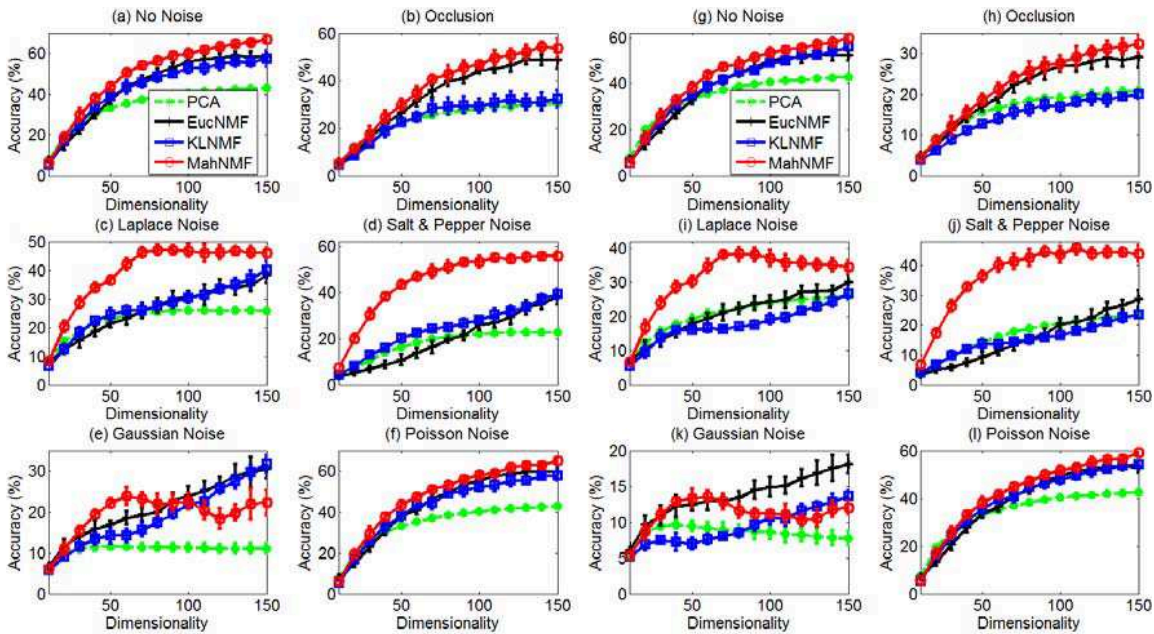}
\caption{Face recognition accuracy versus dimensionalities of PCA, EucNMF, KLNMF, and MahNMF on the Yale B dataset when the training set is contaminated by occlusion, Laplace noise, salt \& pepper noise, Gaussian noise, and Poisson noise. All the algorithms were evaluated under two settings: the test set is clean (a-f) and the test set is contaminated by the same noise as the training set (g-l).}
\label{fig.frc-yaleb}
\end{figure*}

\subsubsection{Yale B Dataset}
The extended Yale face database B \citep{Georghiades2001} contains $16,128$ images of $38$ individuals under $9$ poses and $64$ illumination conditions. All the images are manually aligned and cropped to $32\times 32$ pixels. We simply selected around $64$ near frontal images under different illuminations per individual and reshaped each image into an $1024$-dimensional long vector. We conducted MahNMF, EucNMF and KLNMF on the training set contaminated by five types of outliers with reduced dimensionalities varying from $10$ to $150$. The eigenface obtained by PCA \citep{Hotelling1933} was used as a baseline. Figure \ref{fig.frc-yaleb} gives their average accuracies and standard deviations.

Figure \ref{fig.frc-yaleb} (a) and (g) show that MahNMF outperforms both EucNMF and KLNMF on the Yale B dataset because it is robust to outliers caused by illuminations and shadows. To further study the robustness of MahNMF representation, Figure \ref{fig.frc-yaleb} (c), (d), (i) and (j) show that MahNMF significantly outperforms both EucNMF and KLNMF on the training set contaminated by Laplace noise and Salt \& Pepper noise because MahNMF successfully models such heavy-tailed noises. On the other hand, Figure \ref{fig.frc-yaleb} (e) and (k) show that MahNMF does not perform well when the training set is contaminated by Gaussian noise because it violates the assumption of MahNMF. Similarly, Figure \ref{fig.frc-yaleb} (f) and (l) show that MahNMF is comparable to KLNMF when the training set is contaminated by Poisson noise. Figure \ref{fig.frc-yaleb} (b) and (h) show that MahNMF performs robustly in presence of occlusion because it successfully suppresses the outliers.

\begin{figure*}[ht]
\centering
\includegraphics[width=1.0\linewidth]{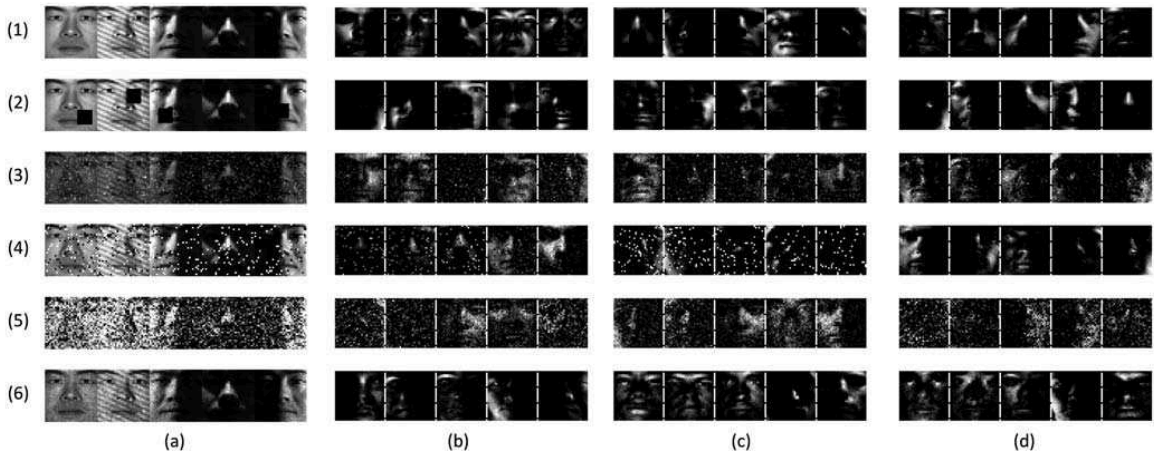}
\caption{Face image examples (column a) of Yale B dataset and the learned basis by KLNMF (column b), EucNMF (column c), and MahNMF (column d) in the absence of noise (1st row) and in the presence of occlusions (2nd row), additive Laplace noise (3rd row), Salt \& Pepper noise (4th row), Gaussian noise (5th row), and multiplicative Poisson noise (6th row).}
\label{fig.basis-yaleb}
\end{figure*}

To further study the effectiveness of MahNMF in data representation, we randomly selected five base vectors from the learned basis by different NMF algorithms when the reduced dimensionality is $50$. Figure \ref{fig.basis-yaleb} compares the base vectors learned by MahNMF with those learned by KLNMF and EucNMF. The first four rows show that MahNMF successfully supresses the occlusion, Laplace noise, and Salt \& Pepper noise while both KLNMF and EucNMF representations are contaminated. It is interesting that MahNMF also suppresses the Poisson noise which confirms the observation in Figure \ref{fig.frc-yaleb}. That is because the Poisson distribution is also heavy-tailed to some extent.

\begin{figure*}[ht]
\centering
\includegraphics[width=1.0\linewidth]{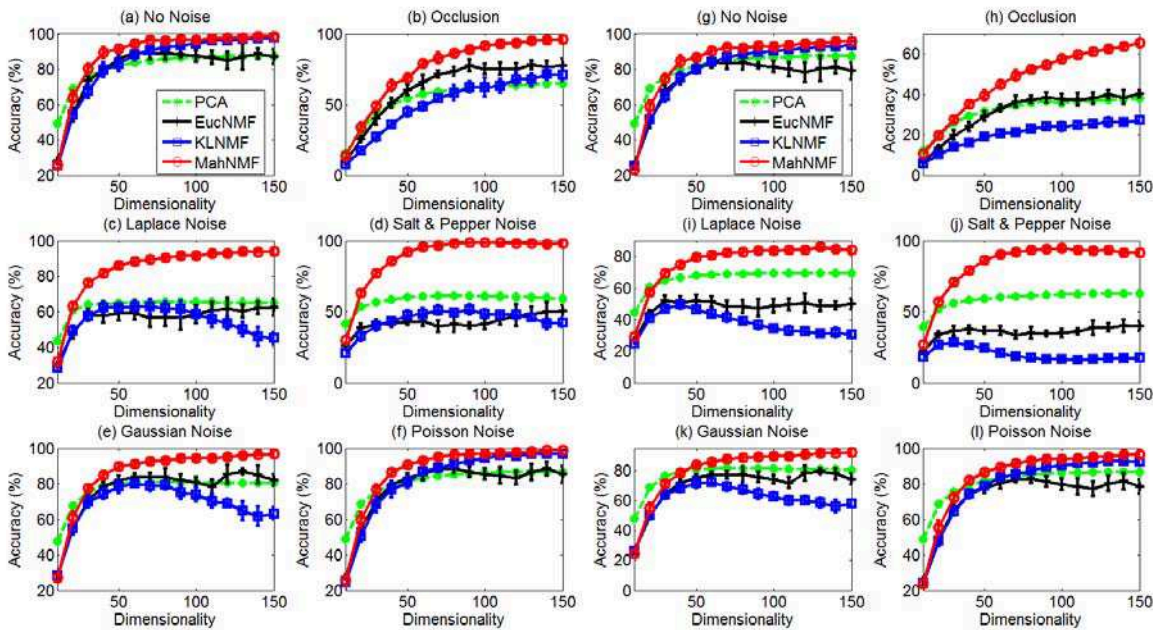}
\caption{Face recognition accuracy versus dimensionalities of PCA, EucNMF, KLNMF, and MahNMF on the PIE dataset when the training set is contaminated by occlusion, Laplace noise, salt \& pepper noise, Gaussian noise, and Poisson noise. All the algorithms were evaluated under two settings: the test set is clean (a-f) and the test set is contaminated by the same noise as the training set (g-l).}
\label{fig.frc-pie}
\end{figure*}

\subsubsection{PIE Dataset}
The CMU PIE face image database \citep{Sim2003} contains $41,368$ images taken from $68$ individuals under $13$ different poses, $43$ different illumination conditions, and $4$ different expressions. All the images have been aligned according to the eye position and cropped to $32\times 32$ pixels. We simply selected $42$ images per individual at Pose $27$ under different light and illumination conditions and reshaped each image into an $1024$-dimensional long vector. Figure \ref{fig.frc-pie} gives the average face recognition accuracies and standard deviations of PCA, EucNMF, KLNMF, and MahNMF representations.

Figure \ref{fig.frc-pie} (a) and (g) shows that MahNMF outperforms EucNMF and its performance is comparable to KLNMF on the PIE dataset. Figures \ref{fig.frc-pie} (b) to (j) show that MahNMF significantly outperforms both EucNMF and KLNMF when the training set is contaminated by occlusion, Laplace noise, and Salt \& Pepper noise. Figure \ref{fig.frc-pie} (a), (d), (g) and (j) show that MahNMF performs almost perfectly on the PIE dataset even when both the training and test sets are seriously contaminated by Salt \& Pepper noise. This is because the PIE face images are mainly contaminated by illumination which can be successfully removed by using the low-rank and sparse representation of MahNMF. For similar reasons, MahNMF outperforms both EucNMF and KLNMF even when the training set is contaminated by Gaussian and Poisson noises (see Figure \ref{fig.frc-pie} (e) to (l)).

\begin{figure*}[ht]
\centering
\includegraphics[width=1.0\linewidth]{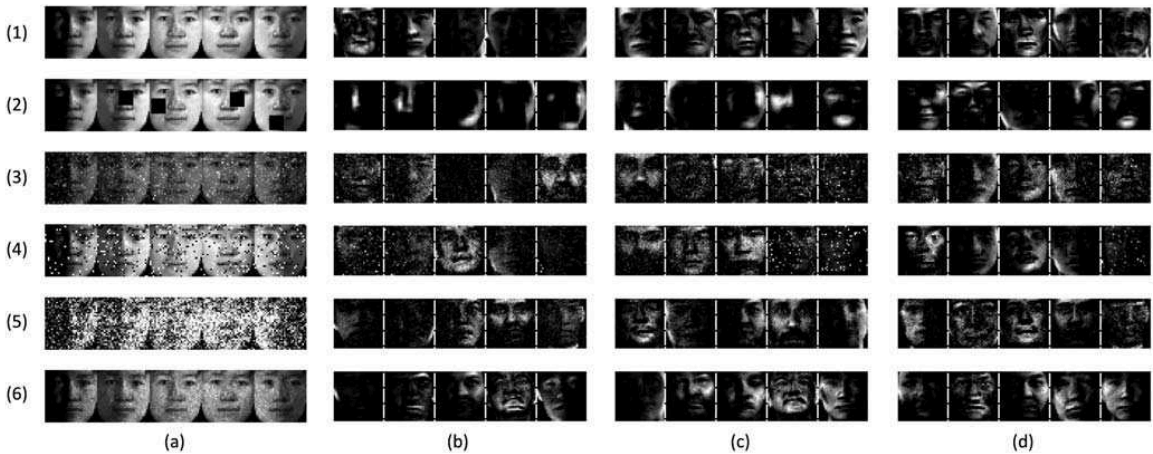}
\caption{Face image examples (column a) of PIE dataset and the learned basis by KLNMF (column b), EucNMF (column c), and MahNMF (column d) in the absence of noise (1st row) and in the presence of occlusions (2nd row), additive Laplace noise (3rd row), Salt \& Pepper noise (4th row), Gaussian noise (5th row), and multiplicative Poisson noise (6th row).}
\label{fig.basis-pie}
\end{figure*}

\begin{figure*}[ht]
\centering
\includegraphics[width=1.0\linewidth]{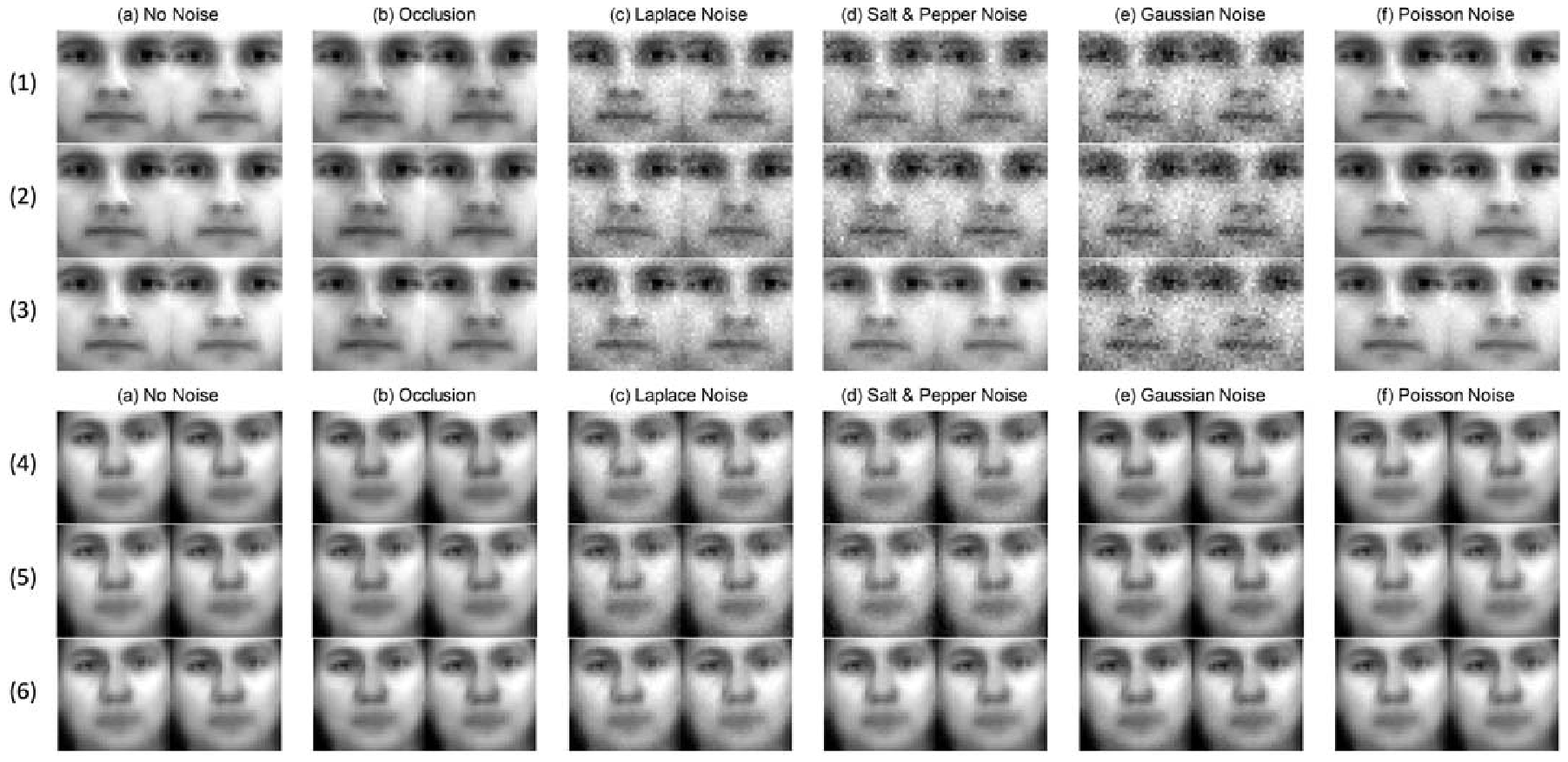}
\caption{Average reconstructed face images of KLNMF (1st and 4th row), EucNMF (2nd and 5th row), and MahNMF (3rd and 6th row) on the Yale B (1st-3rd row) and ORL (4th-6th row) datasets in absence of noise (column a) and in presence of occlusions (column b), additive Laplace noise (column c), Salt \& Pepper noise (column d), Gaussian noise (column e), and multiplicative Poisson noise (column f). The left and right hand images are the averaged reconstructed images on the training set and test set, respectively.}
\label{fig.reconstructions}
\end{figure*}

Figure \ref{fig.basis-pie} gives the randomly selected bases learned by EucNMF, KLNMF and MahNMF when the reduced dimensionality is $50$. It shows that MahNMF successfully suppresses occlusion, Laplace noise, Salt \& Pepper noise, and the illumination in the training set. Therefore, Figure \ref{fig.basis-pie} supports our observations in Figure \ref{fig.frc-pie}.

\begin{table}[ht]
\centering
\caption{The relative errors of the reconstructions by EucNMF, KLNMF, and MahNMF on both Yale B and PIE datasets.}
\resizebox{\textwidth}{!}{%
\begin{tabular}{|c|ccc|ccc|}
\hline
Algorithm	&EucNMF	&KLNMF	&MahNMF	&EucNMF	&KLNMF	&MahNMF \\
\hline
Yale B Dataset	&\multicolumn{3}{c|}{Training Set}	&\multicolumn{3}{c|}{Test Set} \\
\hline
No noise	&$\mathbf{.023\pm.001}$	&$.032\pm.001$	&$.039\pm.002$	&$\mathbf{.055\pm.002}$	&$.058\pm.002$	&$.060\pm.002$ \\
Occlusion	&$\mathbf{.080\pm.004}$	&$.0117\pm.005$	&$.089\pm.004$	&$\mathbf{.095\pm.002}$	&$.120\pm.002$	&$.097\pm.002$ \\
Laplace noise	&$.155\pm.005$	&$.157\pm.004$	&$\mathbf{.127\pm.005}$	&$.140\pm.003$	&$.146\pm.003$	&$\mathbf{.122\pm.002}$ \\
Salt \& Pepper	&$.245\pm.010$	&$.228\pm.013$	&$\mathbf{.082\pm.008}$	&$.133\pm.003$	&$.134\pm.004$	&$\mathbf{.079\pm.003}$ \\
Gaussian noise	&$\mathbf{.241\pm.005}$	&$.263\pm.005$	&$.246\pm.004$	&$.185\pm.004$	&$.194\pm.004$	&$\mathbf{.183\pm.004}$ \\
Poisson noise	&$\mathbf{.028\pm.001}$	&$.034\pm.002$	&$.044\pm.002$	&$\mathbf{.059\pm.002}$	&$.060\pm.002$	&$.062\pm.002$ \\
\hline
PIE Dataset	&\multicolumn{3}{c|}{Training Set}	&\multicolumn{3}{c|}{Test Set} \\
\hline
No noise	&$\mathbf{.010\pm.000}$	&$.009\pm.000$	&$.012\pm.001$	&$.012\pm.000$	&$\mathbf{.012\pm.000}$	&$.014\pm.000$ \\
Occlusion	&$.075\pm.002$	&$.113\pm.003$	&$\mathbf{.063\pm.002}$	&$.070\pm.001$	&$.106\pm.001$	&$\mathbf{.064\pm.001}$ \\
Laplace noise	&$.072\pm.002$	&$.079\pm.003$	&$\mathbf{.046\pm.001}$	&$.056\pm.001$	&$.054\pm.001$	&$\mathbf{.044\pm.001}$ \\
Salt \& Pepper	&$.097\pm.004$	&$.115\pm.003$	&$\mathbf{.021\pm.003}$	&$.052\pm.001$	&$.049\pm.001$	&$\mathbf{.020\pm.001}$ \\
Gaussian noise	&$.024\pm.001$	&$.028\pm.001$	&$\mathbf{.024\pm.001}$	&$.023\pm.000$	&$.024\pm.000$	&$\mathbf{.023\pm.000}$ \\
Poisson noise	&$.013\pm.000$	&$\mathbf{.011\pm.000}$	&$.012\pm.000$	&$.014\pm.000$	&$\mathbf{.013\pm.000}$	&$.014\pm.000$ \\
\hline
\end{tabular}}
\label{table.relative-error}
\end{table}

One may be interested in the reconstruction capacity for the original images when they are contaminated by outliers, e.g., occlusion and noises. Figure \ref{fig.reconstructions} gives the average reconstruction of the face images in both training and test sets of the Yale B and PIE datasets. It shows that MahNMF obtains clearer reconstruction than EucNMF and KLNMF. To further study MahNMF's reconstruction capacity, Table \ref{table.relative-error} compares its relative error with the errors obtained by EucNMF and KLNMF on both Yale B and PIE datasets. The relative error is defined as $\frac{\|X-X^{'}\|_F^2}{\|X\|_F^2}$, wherein $X$ and $X^{'}$ are the original image and reconstructed image, respectively. Here we only compare the relative errors when the reduced dimensionality is $80$. For other reduced dimensionalities, we make similar observations as shown in Table \ref{table.relative-error}. Table \ref{table.relative-error} shows that MahNMF reconstructs the face images better in both the training and test sets when the training set is contaminated by Laplace and Salt \& Pepper noises. Therefore, MahNMF successfully handles the heavy-tailed noise and performs robustly in the presence of outliers.

\subsection{Image Clustering Study}
In this section, we first conduct a simple clustering experiment on face image datasets in the presence of different types of outliers to show its effectiveness in data representation. We then conduct image segmentation experiments to evaluate the clustering effectiveness of MahNMF-\textit{SYM}, since the segmentation problem is intrinsically a clustering problem \citep{Wu1993}.

\subsubsection{Face Image Datasets}
\begin{figure*}[ht]
\centering
\includegraphics[width=1.0\linewidth]{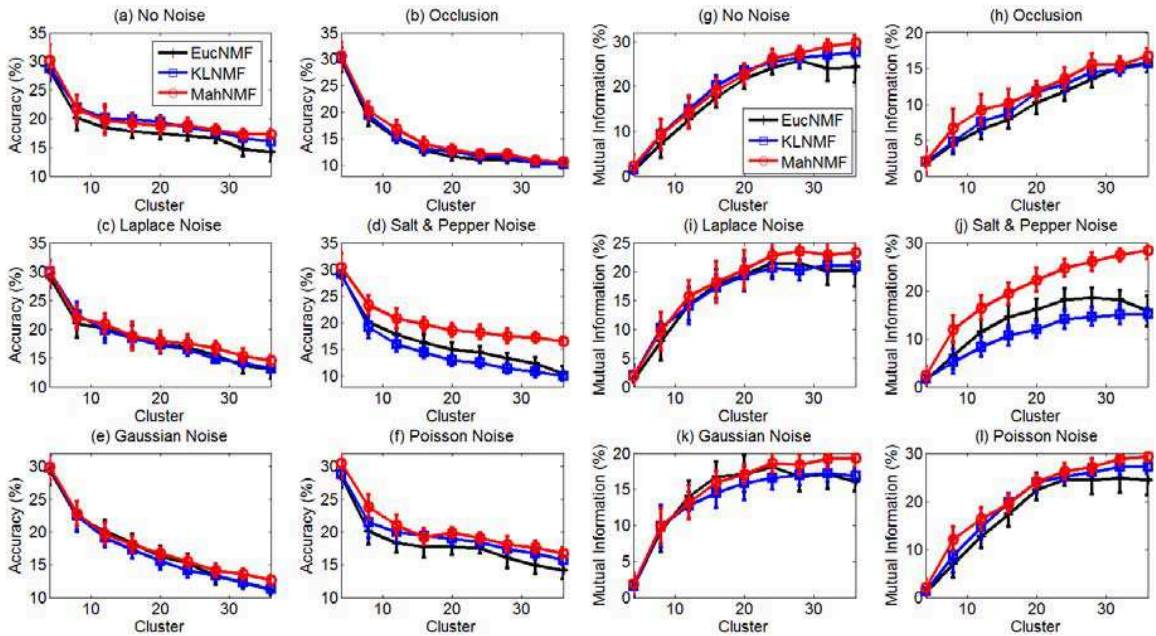}
\caption{Image clustering accuracy (a-f) and mutual information (g-l) versus cluster number of EucNMF, KLNMF, and MahNMF on the Yale B dataset in absence of noise (a \& g) and in presence of occlusion (b \& h), additive Laplace (c \& i), Salt \& Pepper (d \& j), Gaussian noise (e \& k), and Poisson noise (f \& l).}
\label{fig.clu-yaleb}
\end{figure*}

\begin{figure*}[ht]
\centering
\includegraphics[width=1.0\linewidth]{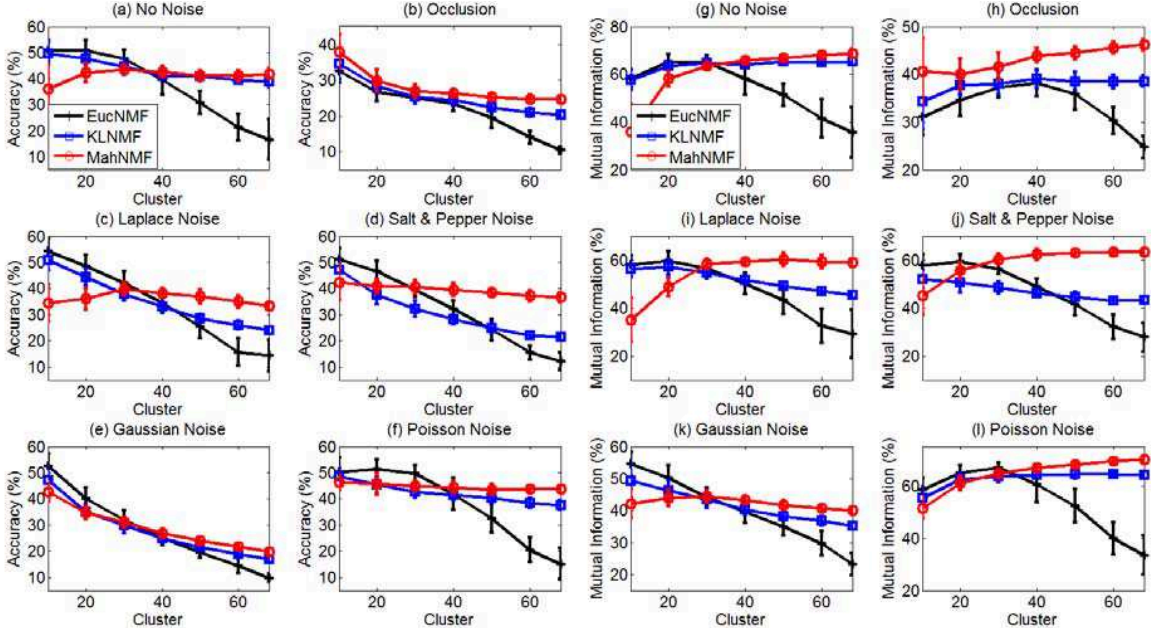}
\caption{Image clustering accuracy (a-f) and mutual information (g-l) versus cluster number of EucNMF, KLNMF, and MahNMF on the PIE dataset in absence of noise (a \& g) and in presence of occlusion (b \& h), additive Laplace (c \& i), Salt \& Pepper (d \& j), Gaussian noise (e \& k), and Poisson noise (f \& l).}
\label{fig.clu-pie}
\end{figure*}

To evaluate the effectiveness and robustness of MahNMF in data representation, we conduct the clustering experiments on both Yale B and ORL datasets. We randomly select $4$ to $36$ individuals from the Yale B dataset and $10$ to $68$ individuals from the ORL dataset to construct the test set $X\in \mathrm{R}_+^{1024\times N}$, wherein $N$ signifies the size of the test set. By factorizing the reweighted $X$ with EucNMF, KLNMF, and MahNMF, we evaluate their clustering performance in terms of both accuracy (AC) and mutual information (MI) \citep{Xu2003}. To eliminate the randomness of individual selection, we repeat this trial $20$ times and report the average AC and MI in Figures \ref{fig.clu-yaleb} and \ref{fig.clu-pie}.

Figures \ref{fig.clu-yaleb} and \ref{fig.clu-pie} show that MahNMF outperforms both EucNMF and KLNMF on the Yale B and PIE datasets even though the test set is contaminated by occlusion, Laplace noise, and salt \& pepper noise. Figure \ref{fig.clu-pie} (e) and (k) show that MahNMF performs better than EucNMF on the PIE dataset in the presence of Gaussian noise because this dataset contains serious illumination outliers and MahNMF can robustly recover the sparse and low rank representation while EucNMF cannot.

\subsubsection{Image Segmentation}
Spectral clustering methods such as normalized cuts (Ncuts, \citep{Shi2000}) have been successfully used in image segmentation. They usually decompose the (normalized) Laplacian matrices by using Eigen decomposition and partition the pixels of image into two parts based on the second eigenvector. By recursively partitioning pixels, Ncuts successfully segment a given image. Recently, Ding \textit{et al}. \citep{Ding2006} have proved that the symmetric EucNMF is equivalent to spectral clustering. We study the effectiveness of our MahNMF-\textit{SYM} algorithm in image segmentation by comparing it with Ncuts.

\begin{figure*}[ht]
\centering
\includegraphics[width=1.0\linewidth]{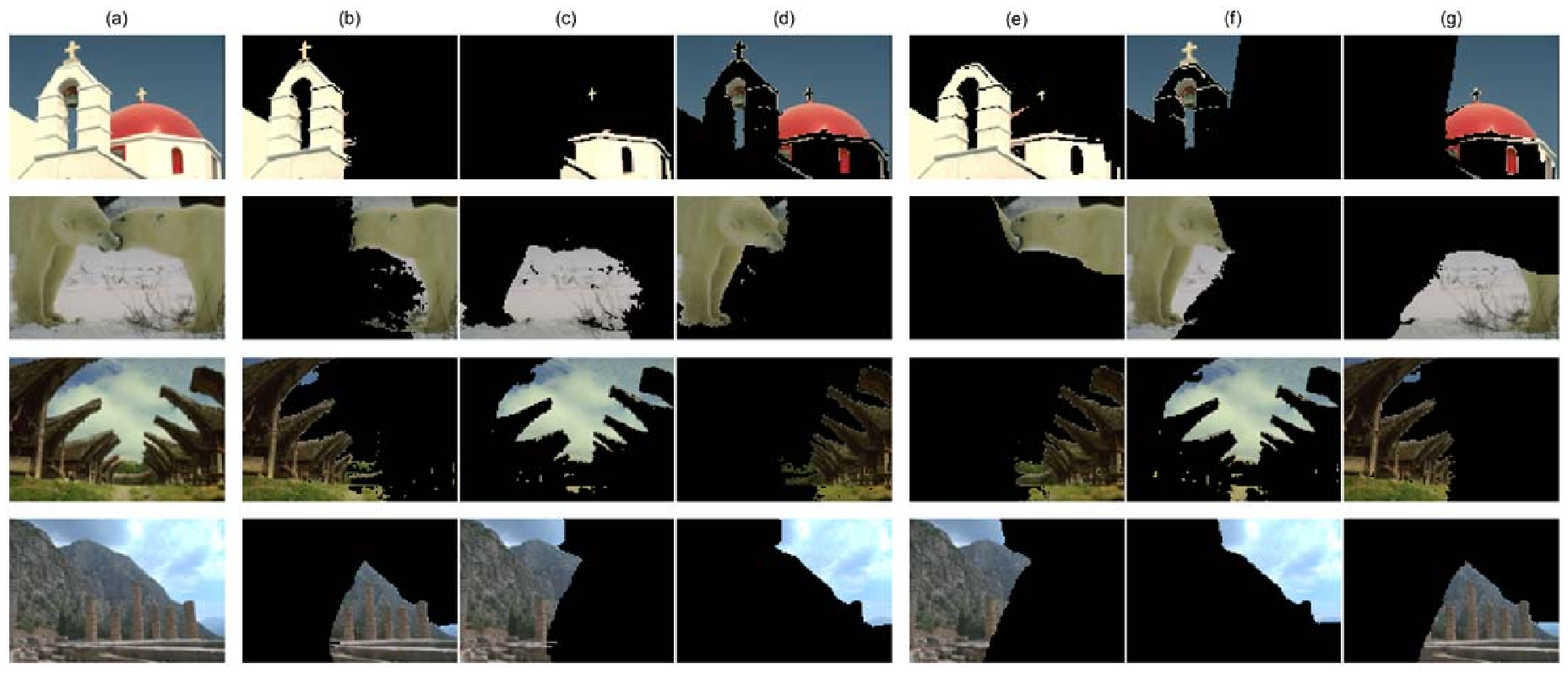}
\caption{Three segments of four example images (a) by using MahNMF-\textit{SYM}+K-means (b to d) and Ncuts (e to g). Each row corresponds to an image.}
\label{fig.three-segs-one}
\end{figure*}

\begin{figure*}[ht]
\centering
\includegraphics[width=1.0\linewidth]{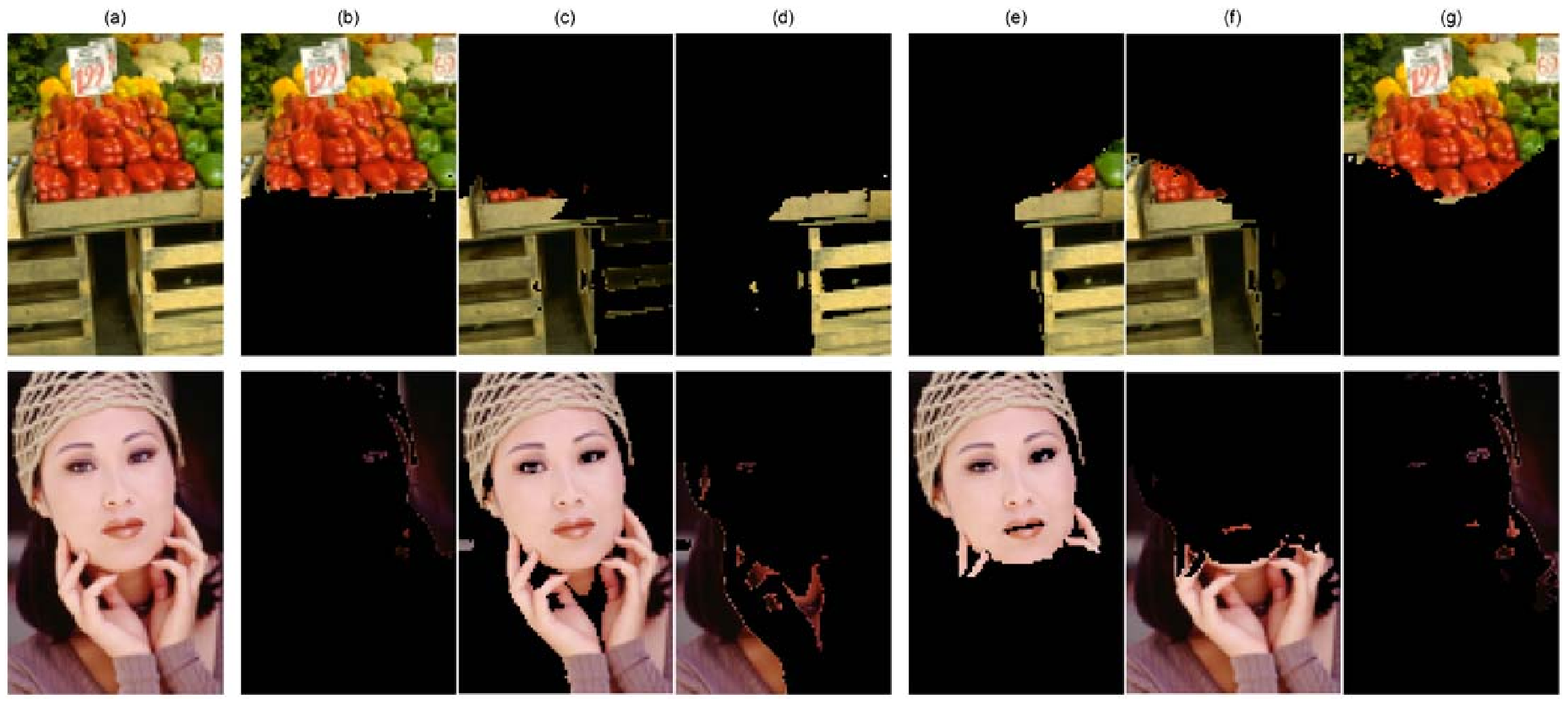}
\caption{Three segments of another two example images (a) by using MahNMF-\textit{SYM}+K-means (b to d) and Ncuts (e to g). Each row corresponds to an image.}
\label{fig.three-segs-two}
\end{figure*}

\begin{figure*}[ht]
\centering
\includegraphics[width=1.0\linewidth]{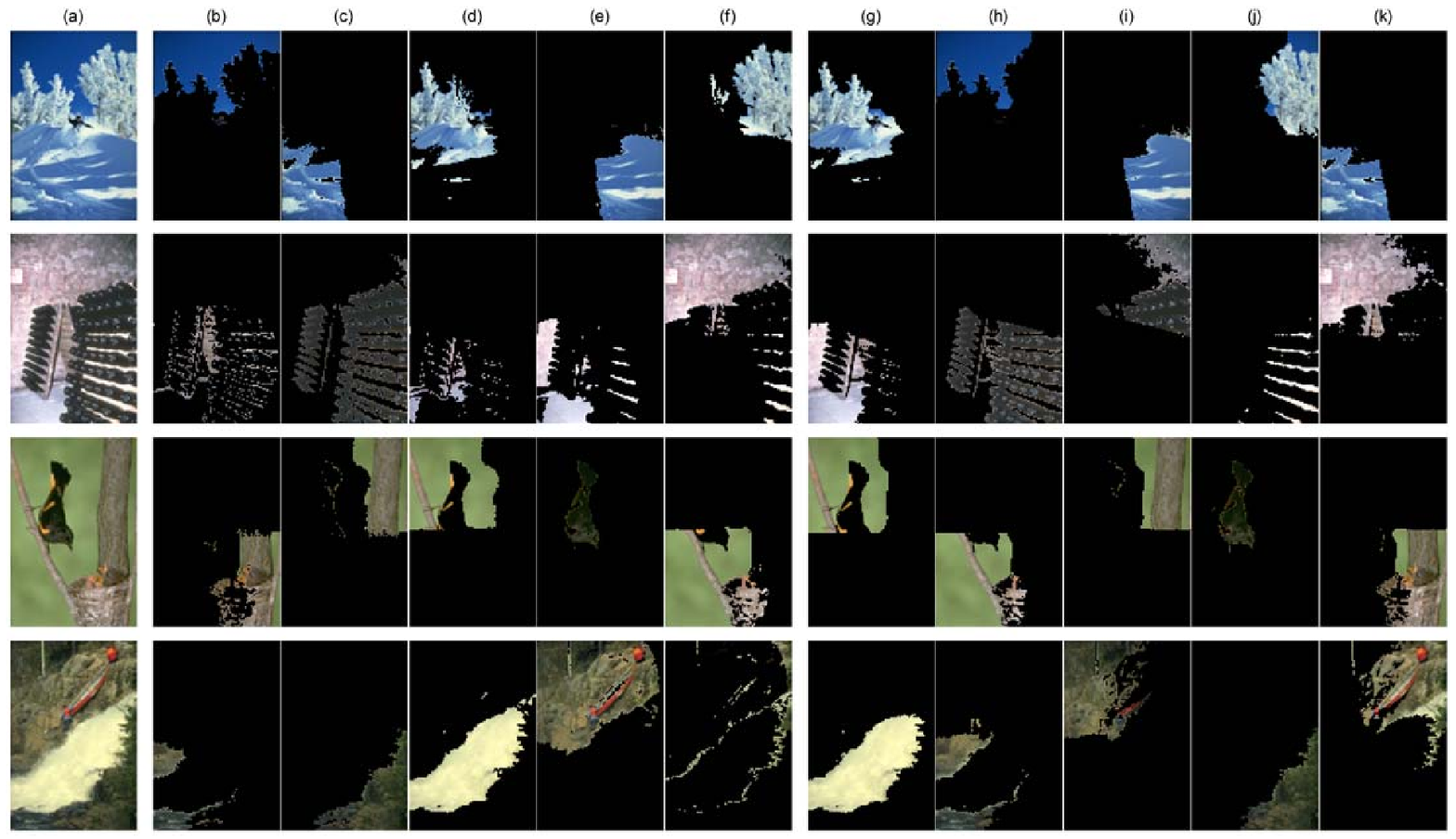}
\caption{Five segments of four example images (a) by using MahNMF-\textit{SYM}+K-means (b to f) and Ncuts (g to k). Each row corresponds to an image.}
\label{fig.five-segs}
\end{figure*}

In this experiment, we compare MahNMF-\textit{SYM} and Ncuts on the Berkeley segmentation dataset \citep{Martin2001}. For a given image, we construct a graph $G$, wherein each node corresponding to a pixel and the edge between node $i$ and node $j$ has the weight $w_{ij}$ defined as the product of a feature similarity term and spatial proximity term. Following \citep{Shi2000}, we set
\begin{equation}
    w_{ij}=e^{-\frac{dF_{(i,j)}^2}{\delta_F^2}}\times \left\{\begin{array}{c@{\;}l} e^{-\frac{dL_{(i,j)}^2}{\delta_L^2}}, &dL_{(i,j)}\le r \\0, &otherwise \end{array}\right., \label{eq5.1}
\end{equation}
where $dF_{(i,j)}=\|F(i)-F(j)\|_{l_2}$ and $F(i)$ denotes the brightness of node $i$, $dL_{(i,j)}=\|L(i)-L(j)\|_{l_2}$ and $L(i)$ denote the spatial location of node $i$. The parameter $r$ is used to suppress the correlation between two pixels that are relatively far from each other. Ncuts partition $G$ based on the second eigenvector of the normalized Laplacian matrix $L=I-D^{-\frac{1}{2}}WD^{-\frac{1}{2}}$, wherein $D$ is a diagonal matrix with $D_{ii}=\sum_{j=1}^N W_{ij}$ and $N$ is the total pixel number. Similarly, we factorize the normalized similarity matrix $D^{-\frac{1}{2}}WD^{-\frac{1}{2}}\approx HH^T$ by using the proposed MahNMF-\textit{SYM} algorithm followed by clustering $H$ with K-means to obtain labels for all the segments. The reduced dimensionality is set to $5$ in MahNMF-\textit{SYM} and the segments number is set to $3$ and $5$, respectively. Figures \ref{fig.three-segs-one}, \ref{fig.three-segs-two}, and \ref{fig.five-segs} give segmentation results for ten example images.

\begin{figure*}[ht]
\centering
\includegraphics[width=1.0\linewidth]{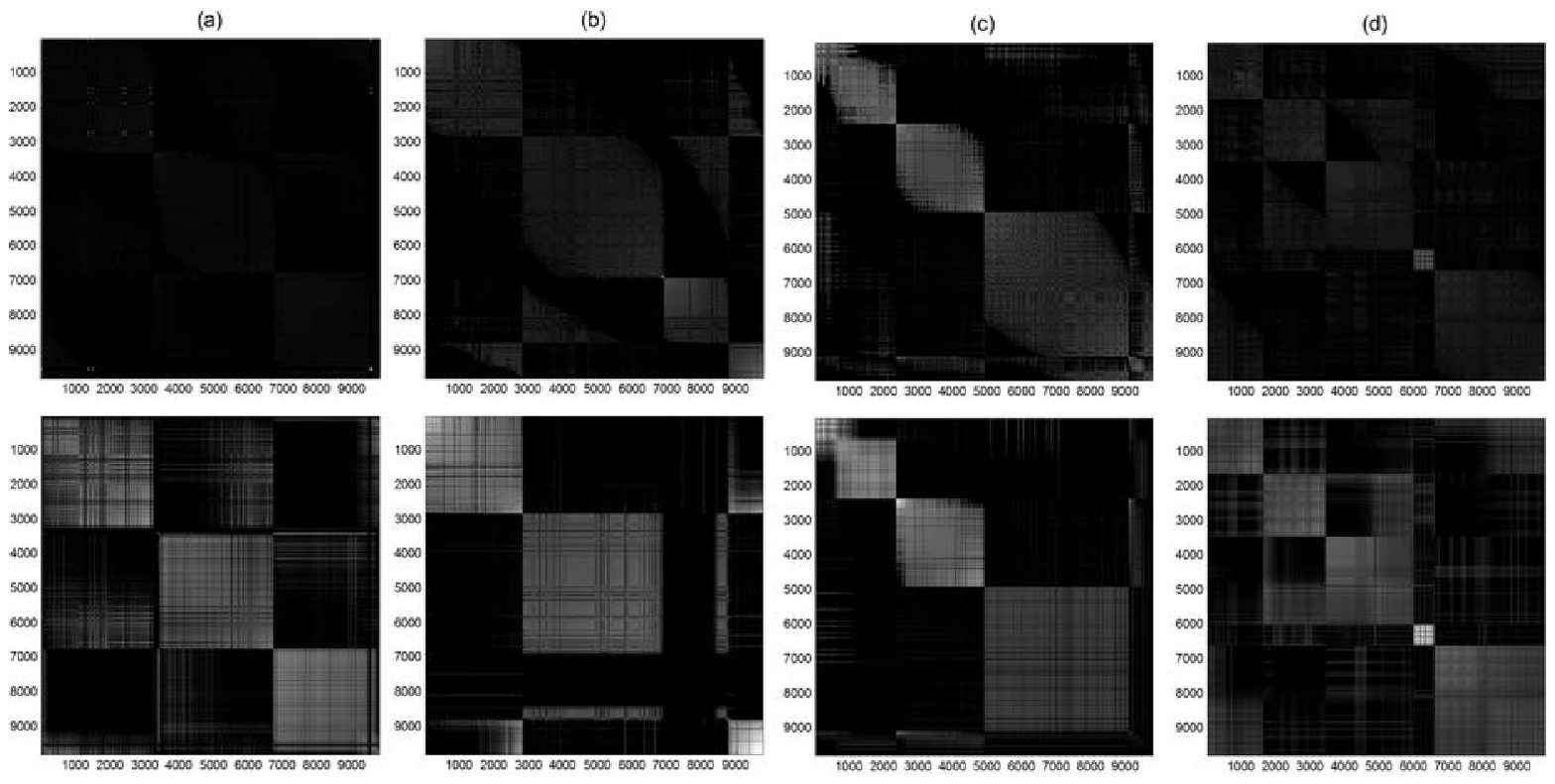}
\caption{Five segments of four example images (a) by using MahNMF-\textit{SYM}+K-means (b to f) and Ncuts (g to k). Each row corresponds to an image.}
\label{fig.correlations}
\end{figure*}

Figures \ref{fig.three-segs-one}, \ref{fig.three-segs-two}, and \ref{fig.five-segs} show that MahNMF-\textit{SYM}+K-means successfully separates the objects in these example images and its performance is mostly comparable with Ncuts. In some cases, e.g., the second rows in Figure \ref{fig.three-segs-one}, \ref{fig.three-segs-two}, and \ref{fig.five-segs}, MahNMF-\textit{SYM}+K-means outperforms Ncuts. Figure \ref{fig.correlations} shows the normalized similarity matrices, i.e., $D^{-\frac{1}{2}}WD^{-\frac{1}{2}}$, for four example images and their low-rank approximations, i.e., $HH^T$. From Figure \ref{fig.correlations} (a) and (b), we can find that MahNMF-\textit{SYM} clearly distinguishes the three classes of correlations between pixels of images of the first and second rows of Figure \ref{fig.three-segs-one}. Therefore, low-rank representation helps K-means to correctly segment these images into three parts. Similarly, according to Figure \ref{fig.correlations} (c) and (d), MahNMF-\textit{SYM} successfully finds the correlations between pixels of images of the second and last rows of Figure \ref{fig.five-segs}, and thus the learned low-rank representation helps K-means to segment these images into five parts. It means that the MahNMF-\textit{SYM}+K-means method performs well for image segmentation.

Note that the parameters $\delta_F$ and $\delta_L$ in \eqref{eq5.1} should be carefully selected \citep{Nascimento2011}. In this experiment, we normalized both $dF$ and $dL$ to $[0,1]$, and simply set $\delta_F=.3$ and $\delta_L=.7$. Another critical parameter $r$ was set to $r=med(dL)$. Because this experiment aims to study the effectiveness of MahNMF-\textit{SYM} in image segmentation, we deduce tuning these parameters to future work.

\subsection{Sparse and Low-rank Decomposition}
In this section, we conduct background subtraction and shadow/illumination removal experiments to evaluate the sparse and low-rank decomposition capability of MahNMF by comparing it with robust principal component analysis (RPCA\footnote{http://perception.csl.uiuc.edu/matrix-rank/sample\_code.html}, \citep{Candes2011}) and GoDec\footnote{https://sites.google.com/site/godecomposition/code} \citep{Zhou2011}.

\begin{figure*}[ht]
\centering
\includegraphics[width=1.0\linewidth]{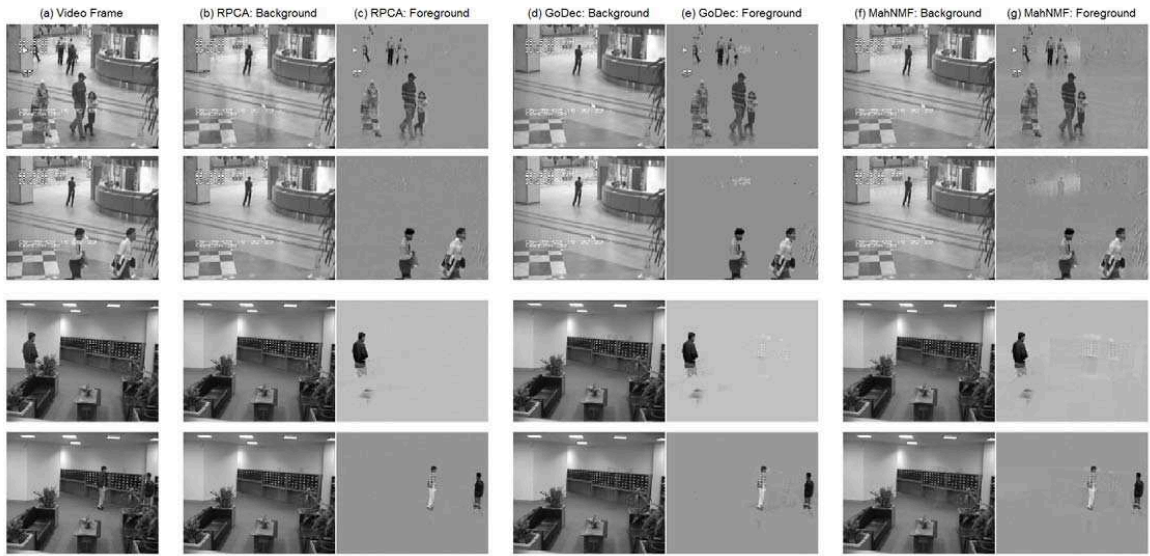}
\caption{Video frame (column a), background and foreground subtracted by RPCA (column b and c), GoDec (column d and e), and MahNMF (column f and g) on `Hall' (first two rows) and `Lobby' (last two rows) video sequences.}
\label{fig.bk200-hall-lobby}
\end{figure*}

\begin{figure*}[ht]
\centering
\includegraphics[width=1.0\linewidth]{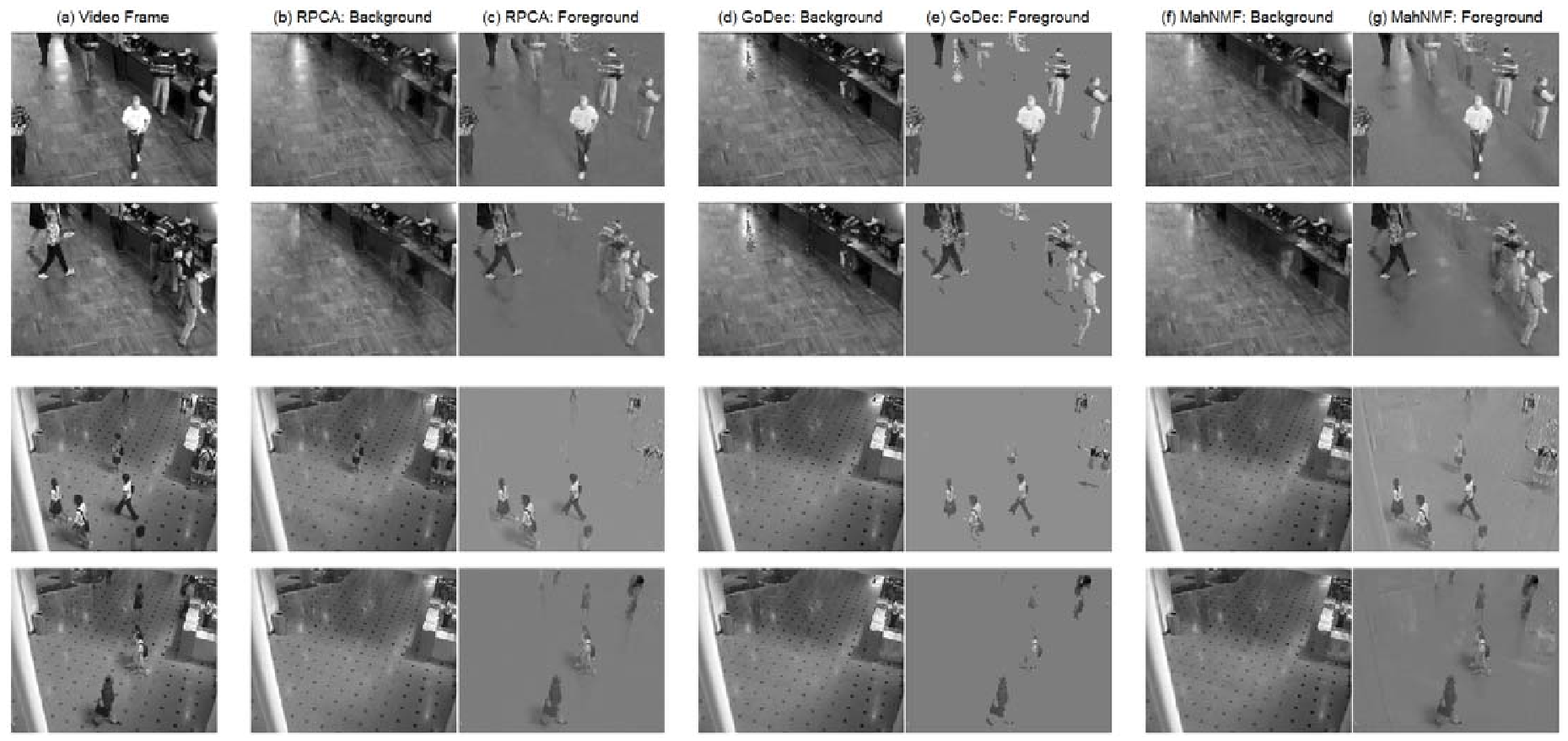}
\caption{Video frame (column a), background and foreground subtracted by RPCA (column b and c), GoDec (column d and e), and MahNMF (column f and g) on `Bootstrap' (first two rows) and `Shopping Mall' (last two rows) video sequences.}
\label{fig.bk200-bootstrap-shoppingmall}
\end{figure*}

\subsubsection{Background Subtraction}
In video surveillance, background modeling is a challenging task that models the background and detects moving objects in the foreground \citep{Cheng2011}. The background variation can be approximated by low-rank representation because video frames may share the same background. Moreover, the foreground objects, such as walkers or cars, occupy only few image pixels and thus can be considered as sparse noise. As discussed in Section I, MahNMF naturally reveals the low-rank approximation of background and stores the foreground moving objects in noise. In this experiment, we evaluate its capability in background subtraction on four surveillance videos\footnote{http://perception.i2r.a-star.edu.sg/bk\_model/bk\_index.html} \citep{Li2004} including `Hall', `Lobby', `Bootstrap', and `Shopping Mall', whose resolutions are $144\times 176$, $128\times 160$, $120\times 160$, and $256\times 320$, respectively. Similar to \citep{Candes2011} and \citep{Zhou2011}, we select $200$ frames from each video and resize each frame to a long vector to construct the data matrix $X$. We set the low rank of MahNMF to $2$ and keep the parameter settings of RPCA and GoDec consistent with those given in their demos. Figures \ref{fig.bk200-hall-lobby} and \ref{fig.bk200-bootstrap-shoppingmall} give the background and foreground of the first two and last two videos ($2$ frames per video). These figures show that MahNMF successfully separates the background and foreground without losing detail. The results obtained by MahNMF are comparable to those obtained by RPCA and GoDec.

\begin{figure*}[ht]
\centering
\includegraphics[width=1.0\linewidth]{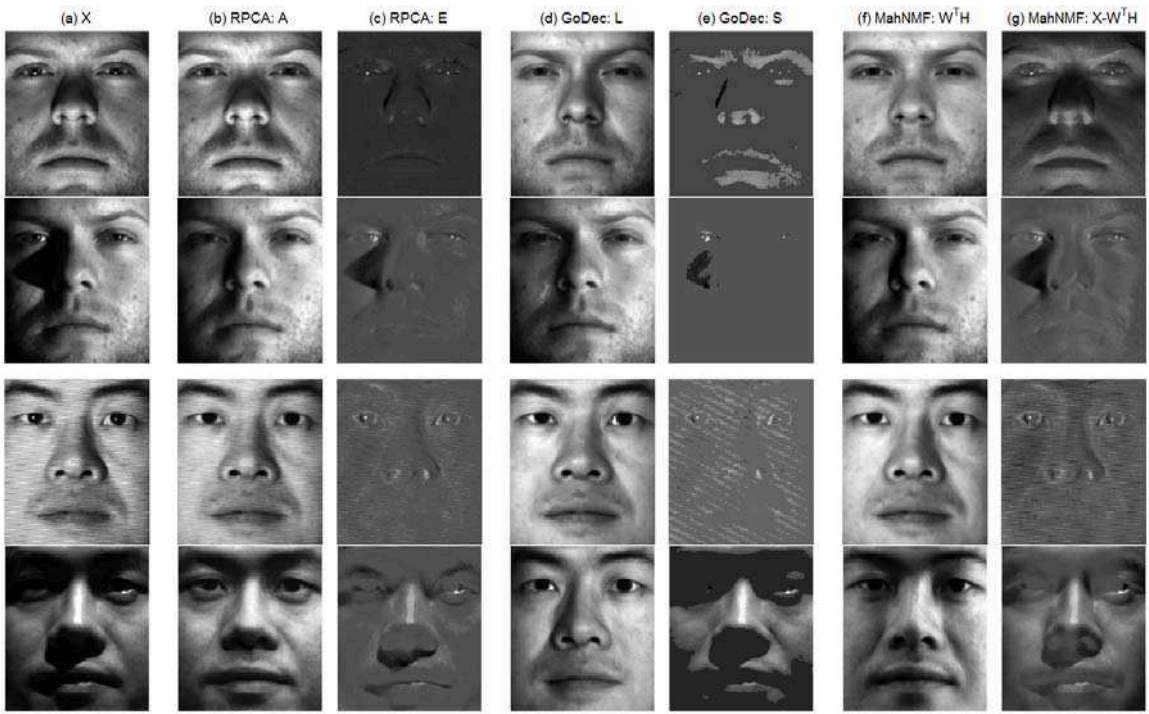}
\caption{The shadow/illumination removal results obtained by RPCA ($X=A+E$), GoDec ($X=L+S+E$), and MahNMF ($X\approx W^T H$) on face images of the first two individuals taken from the extended Yale B dataset, where the top two rows come from the first individual and the bottom two rows come from the second individual.}
\label{fig.face-1-2}
\end{figure*}

\begin{figure*}[ht]
\centering
\includegraphics[width=1.0\linewidth]{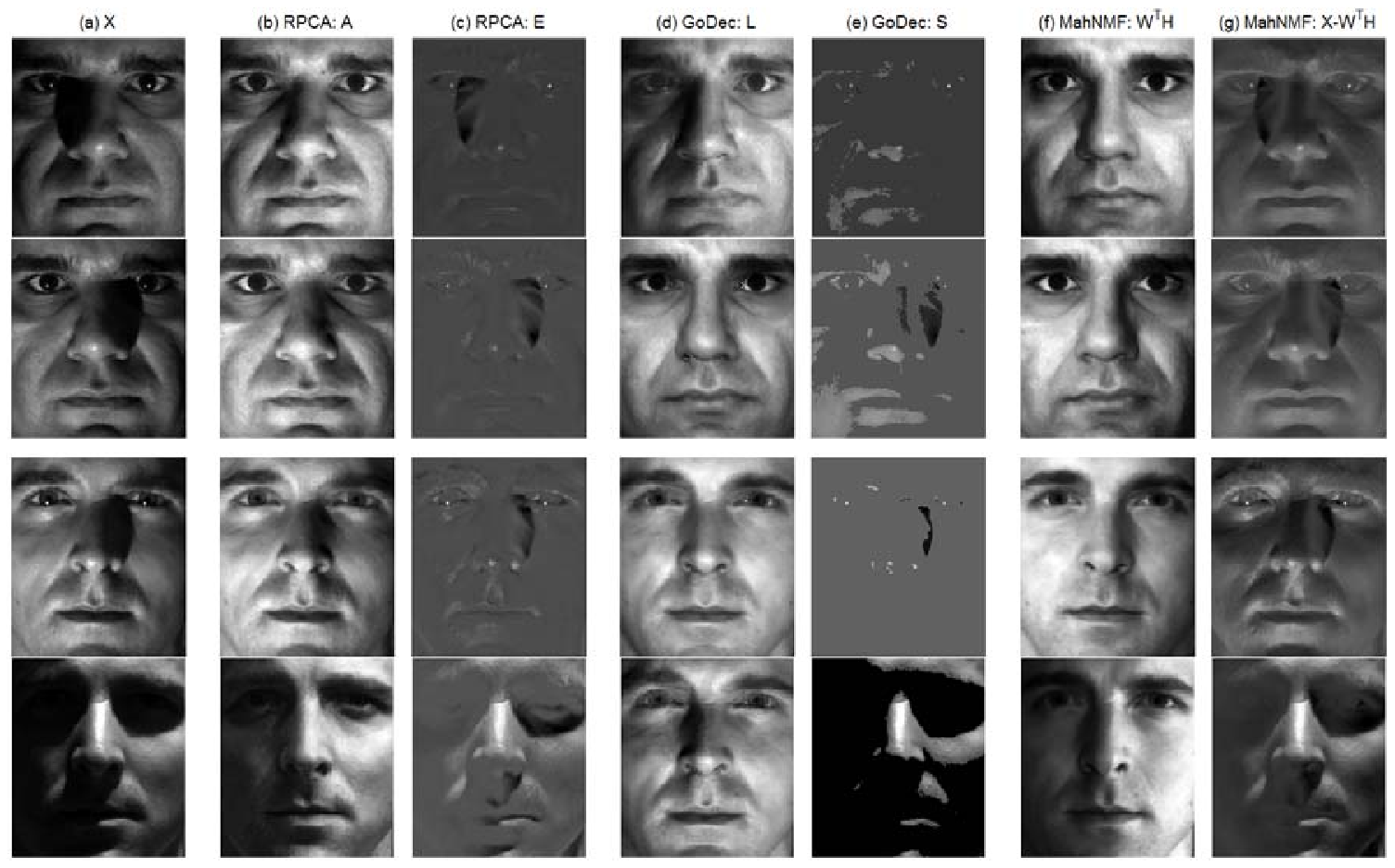}
\caption{The shadow/illumination removal results obtained by RPCA ($X=A+E$), GoDec ($X=L+S+E$), and MahNMF ($X\approx W^T H$) on face images of the first two individuals taken from the extended Yale B dataset, where the top two rows come from the third individual and the bottom two rows come from the fourth individual.}
\label{fig.face-8-10}
\end{figure*}

\subsubsection{Shadow/Illumination Removal}
According to the face recognition results in Section V.B, both shadow and illumination pull down the image quality and thus reduce the accuracy of many methods such as principal component analysis (PCA, \citep{Hotelling1933}). However, MahNMF performs well on both Yale B and PIE datasets which are seriously contaminated by shadow and illumination. That is because MahNMF robustly estimates the low-rank representation of the face images. In this experiment, we further study the capability of MahNMF in shadow/illumination removal by comparing with RPCA and GoDec.

We conducted MahNMF, RPCA, and GoDec on the images taken from single individual of the extended Yale B dataset\footnote{http://cvc.yale.edu/projects/yalefacesB/yalefacesB.html} which contains around $64$ frontal face images taken from $38$ individuals. By reshaping the $192\times 168$ pixels of each image into a long vector, we got a $32764\times 64$-dimensional data matrix. We set the low rank of MahNMF and GoDec to $3$ and kept other settings consistent with those in the background subtraction experiment. Figures \ref{fig.face-1-2} and \ref{fig.face-8-10} give the shadow/illumination removal results for four individuals. It shows that MahNMF successfully removes both shadow and illumination of these images. Such observation confirms those results obtained from Section V.B. From Figures \ref{fig.face-1-2} and \ref{fig.face-8-10}, we can see that the performance of MahNMF is comparable to RPCA and GoDec.

\subsection{Multi-View Learning}
In several computer vision tasks, data sets often inherently involve many types of features such as pixels, gradient-based features, color-based features, and surrounding text, which represent the same image from different views. Many computer vision tasks such as image retrieval and image annotation have proven to be beneficial from these multiple views. One of the most important problems is how to learn a latent representation of the data to leverage the information shared by the multiple views. To this end, Jia \textit{et al}. \citep{Jia2010} proposed factorized space with a structured sparsity (FLSS) algorithm that learns a latent space to factorize the information contained in multiple views into shared and private parts. Given a data set $X\in R^{M\times N}$, where the features are composed of $V$ views, FLSS learns to find the dictionary $D$ and the coefficients $H$ by optimizing
\begin{equation}
    \min_{D,H} \frac{1}{N} \|X-D^T H\|_F^2+\lambda\sum_{v=1}^V \|D_{[\rho_v]}^T\|_{1,\infty}+\gamma \|H\|_{1,\infty}, \label{eq5.2}
\end{equation}
where $\lambda>0$ and $\gamma>0$ are the weights of the group sparsity over $D$ and $H$, respectively, and $\rho_v$ is the index for the $v$-th view. According to \citep{Jia2010}, the parameters in \eqref{eq5.2} were set to $\lambda=.01$ and $\gamma=.01$ in this experiment. In contrast to \citep{Jia2010}, Kim \textit{et al}. \citep{Kim2012} proposed a group sparse NMF to learn a low-rank representation of the multi-view data by incorporating the $l_{1,p}$ norm over the basis matrix into EucNMF's loss function, i.e.,
\begin{equation}
    \min_{W\ge 0,H\ge 0} \frac{1}{2} \|X-W^T H\|_F^2+\alpha \sum_{v=1}^V \|W_{[\rho_v]}^T\|_{1,p}+\frac{\beta}{2}\|H\|_F^2, \label{eq5.3}
\end{equation}
where $p=2,\infty$ and $\alpha$ and $\beta$ are the weights of the group sparsity of $W$ and Tikhonove regularization over $H$, respectively. Here we term this algorithm EucNMF-\textit{GS} and set its parameters as $\alpha=10^{-2}$ and $\beta=10^{-4}$ according to \citep{Kim2012}. Although EucNMF-\textit{GS} achieves great success in multilingual text analysis, it is not robust in computer vision applications because the Euclidean distance-based model fails to handle the heavy-tailed noise contained in some features such as SIFT \citep{Jia2011}. The proposed MahNMF-\textit{GS} overcomes this problem and performs robustly in multi-view learning. In this experiment, we use the MahNMF-\textit{GS} model defined in \eqref{eq4.9} with the parameters $\gamma_W$ setting to $1\%$ of the group sparsity over initial $W^{[\rho]}$, wherein $\rho\in\{1,...,V\}$. To keep consistent with \citep{Jia2010}, we set $p=\infty$ in both \eqref{eq5.3} and \eqref{eq4.9}.

\begin{figure*}[ht]
\centering
\includegraphics[width=1.0\linewidth]{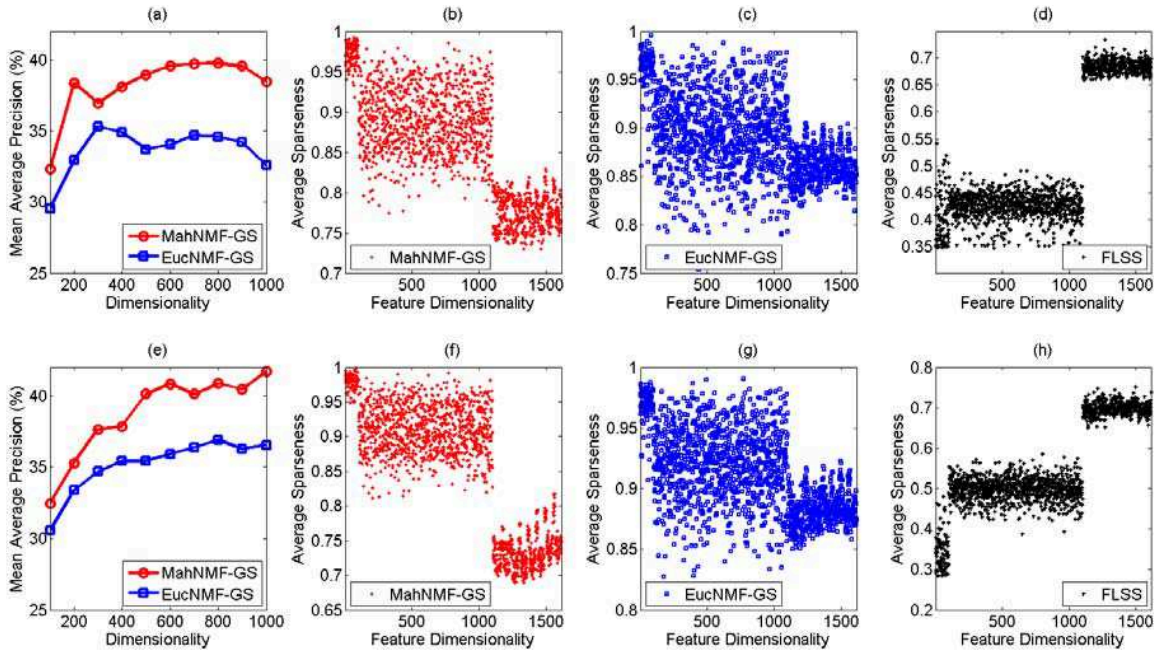}
\caption{The mean average precisions (mAP) and sparsity patterns of the latent representation learned by MahNMF-\textit{GS} and EucNMF-\textit{GS} on both VoC Pascal 07 (the 1st row) and Mir Flickr (the 2nd row) datasets.}
\label{fig.multi-view}
\end{figure*}

We evaluate the effectiveness of MahNMF-\textit{GS} in multi-view learning by comparing it with FLSS and EucNMF-\textit{GS} on two challenging datasets, i.e., VOC Pascal 07 \citep{Everingham2007} and Mirflickr \citep{Huiskes2008} , which contain $10,000$ and $25,000$ natural images collected from $20$ and $38$ classes of objects, respectively. Both datasets contain sixteen types of features from which we select two types of gradient-based features including 1000-dimensional ``DenseSift'' and $512$-dimensional ``Gist'' and one type of color related feature, i.e., $100$-dimensional ``DenseHue'', in training. The training set is constructed by selecting half the images and the remaining images make up the test set. We vary the reduced dimensionality from $100$ to $1,000$ for each dataset and obtain the latent spaces by using MahNMF-\textit{GS}, FLSS, and EucNMF-\textit{GS}. In the test stage, both the training and test samples are projected into the latent space and a SVM classifier for each class of object is constructed. Since this experiment aims to compare the different latent spaces learned by MahNMF-\textit{GS}, FLSS, and EucNMF-\textit{GS}, all SVM classifiers use linear kernel. Based on the constructed classifiers, average precision (AP) is calculated for each class and the mean average precision (mAP) is calculated for evaluation.

\begin{table}[ht]
\centering
\caption{Best mAP Values and the Corresponding Reduced Dimensionalities of MahNMF-\textit{GS}, EucNMF-\textit{GS}, and FLSS on Both VOC Pascal07 and Mir Flickr Datasets.}
\begin{tabular}{|c|cc|cc|cc|}
\hline
Algorithm & \multicolumn{2}{c|}{MahNMF-\textit{GS}} & \multicolumn{2}{c|}{EucNMF-\textit{GS}} & \multicolumn{2}{c|}{FLSS} \\
\hline
Dataset	& mAP	& rDim	& mAP	& rDim	& mAP	& rDim \\
VOC Pascal 07	&39.76\%	&800	&35.29\%	&300	&32.15\%	&143\\
Mir Flickr	&41.69\%	&1000	&36.89\%	&800	&32.04\%	&106\\
\hline
\multicolumn{7}{l}{rDim: reduced dimensionality.}
\end{tabular}
\label{table.multi-view}
\end{table}

Figure \ref{fig.multi-view} presents the mAP versus dimensionalities for MahNMF-\textit{GS} and EucNMF-\textit{GS}. It shows that MahNMF-\textit{GS} significantly outperforms EucNMF-\textit{GS} on both datasets. To study the latent spaces learned by different algorithms, we presented the average sparseness of each column of the learned basis matrices in the second and third columns of Figure \ref{fig.multi-view}. According to \citep{Hoyer2004}, the sparseness of an $N$-dimensional vector $\vec{x}$ is defined as
\begin{equation}
    SPR(\vec{x})=\frac{\sqrt{N}-{\|\vec{x}\|_{l_1}}/{\|\vec{x}\|_{l_2}}}{\sqrt{N}-1}. \label{eq5.4}
\end{equation}
According to \eqref{eq5.4}, the fewer non-zeros the vector contains, the larger its sparseness. In Figures \ref{fig.multi-view} (b), (c), (e) and (f), the $x$-axis denotes the feature dimensionalities concatenating all the views and the $y$-axis denotes the average sparseness over different reduced dimensionalities. Figures \ref{fig.multi-view} (b) and (e) show that the sparsity pattern of the basis learned by MahNMF-\textit{GS} for each view is similar, while the sparsity patterns for different views are different from one another. This implies that MahNMF-\textit{GS} successfully learns the private information for different views and thus works well in multi-view learning while EucNMF-\textit{GS} does not work well. Table \ref{table.multi-view} shows the best mAP values and the corresponding dimensionalities of the latent subspaces learned by MahNMF-\textit{GS}, EucNMF-\textit{GS} and FLSS, respectively. It shows that MahNMF-\textit{GS} outperforms FLSS on both VOC Pascal 07 and Mir Flickr datasets because FLSS is not designed for classification.

\section{Conclusion}
This paper presents a general MahNMF framework to model the heavy-tailed Laplacian noise by minimizing the Manhattan distance between a data matrix and its low-rank approximation. Compared to traditional NMF, MahNMF is much more robust to outliers including both occlusions and several types of noises, and thus performs well in both classification and clustering. Since MahNMF naturally takes into account prior knowledge about the underlying low-rank structure of data and sparse structure of noise, it robustly recovers the low-rank and sparse parts of a non-negative matrix, and its performance is comparable to robust principal component analysis (RPCA) and GoDec in background subtraction and illumination/shadow modeling. While RPCA and GoDec are suitable for matrix completion, MahNMF is well-suited for data representation with the non-negativity property of data kept. The MahNMF problem is difficult to solve because the objective function is neither convex nor smooth. This paper proposes two fast optimization methods including rank-one residual iteration (RRI) and Nesterov's smoothing method for optimizing MahNMF. RRI successively updates each variable in MahNMF in a closed form solution and thus converges fast. However, its time complexity is high which makes RRI unsuitable for scaling to large scale matrices. The proposed Nesterov smoothing method overcomes this deficiency by optimizing MahNMF with an optimal gradient method on a smartly smoothed approximation function. By setting the smoothness parameter inversely proportional to the iteration number, the Nesterov smoothing method iteratively improves approximation accuracy and converges to an approximate solution of MahNMF. Under the MahNMF framework, we develop box constrained MahNMF, manifold regularized MahNMF, group sparse MahNMF, and elastic net inducing MahNMF and apply the proposed RRI and Nesterov's smoothing method to optimize them. Inspired by spectral clustering, we further develop symmetric MahNMF for image segmentation and discussed its equivalence to normalized cuts (Ncuts). Experimental results on several computer vision problems show that these MahNMF variants are comparable to traditional methods.

%% Acknowledgements should go at the end, before appendices and references
%
%\acks{We would like to acknowledge support for this project
%from the National Science Foundation (NSF grant IIS-9988642)
%and the Multidisciplinary Research Program of the Department
%of Defense (MURI N00014-00-1-0637). }
%
%% Manual newpage inserted to improve layout of sample file - not
%% needed in general before appendices/bibliography.
%
%\newpage

\vskip 0.2in
\bibliography{BibMahNMF}

\end{document}